\documentclass[a4paper,fleqn]{cas-sc}

\usepackage{amssymb}
\usepackage{graphicx}       
\usepackage{hyperref}
\usepackage{subcaption}    
\usepackage{amsmath}       
\usepackage{multirow}
\usepackage{color}
\usepackage{float}

\usepackage[authoryear,longnamesfirst]{natbib}

\def\tsc#1{\csdef{#1}{\textsc{\lowercase{#1}}\xspace}}
\tsc{WGM}
\tsc{QE}

\begin{document}
\let\WriteBookmarks\relax
\def\floatpagepagefraction{1}
\def\textpagefraction{.001}

\shorttitle{}    

\shortauthors{}  

\title[mode = title]{Beyond Mamba: Enhancing State-space Models with Deformable Dilated Convolutions for Multi-scale Traffic Object Detection}  



%

\author{Jun Li}

\cormark[1]

\ead{lijuncst@njnu.edu.cn}

\credit{Conceptualization of this study, Methodology, Software}

\author{Yingying Shi}
\credit{Methodology}
\author{Zhixuan Ruan}
\author{Nan Guo}
\author{Jianhua Xu}


\affiliation{organization={School of Computer and Electronic Information, Nanjing Normal University},
            postcode={210023}, 
            country={China}}

\cortext[1]{Corresponding author}



\begin{abstract}
In a real-world traffic scenario, varying-scale objects are usually distributed in a cluttered background, which poses great challenges to accurate detection. Although current Mamba-based methods can efficiently model long-range dependencies, they still struggle to capture small objects with abundant local details, which hinders joint modeling of local structures and global semantics. Moreover, state-space models exhibit limited hierarchical feature representation and weak cross-scale interaction due to flat sequential modeling and insufficient spatial inductive biases, leading to sub-optimal performance in complex scenes. To address these issues, we propose a Mamba with Deformable Dilated Convolutions Network (MDDCNet) for accurate traffic object detection in this study. In MDDCNet, a well-designed hybrid backbone with successive Multi-Scale Deformable Dilated Convolution (MSDDC) blocks and Mamba blocks enables hierarchical feature representation from local details to global semantics. Meanwhile, a Channel-Enhanced Feed-Forward Network (CE-FFN) is further devised to overcome the limited channel interaction capability of conventional feed-forward networks, whilst a Mamba-based Attention-Aggregating Feature Pyramid Network ($A^2$FPN) is constructed to achieve enhanced multi-scale feature fusion and interaction. Extensive experimental results on public benchmark and real-world datasets demonstrate the superiority of our method over various advanced detectors. The code is available at \url{https://github.com/Bettermea/MDDCNet}.
\end{abstract}


\begin{highlights}
\item We propose a Mamba with Deformable Dilated Convolution detection Network (MDDCNet) for object detection in complex traffic scenarios. MDDCNet features a well-designed hierarchical hybrid backbone containing successive MDDCNet and Mamba blocks.
\item Within our MDDCNet and Mamba blocks, we develop a channel-enhanced feedforward network (CE-FFN) to enable adaptive channel reweighting while preserving strong local spatial modeling.
\item For detection neck, we propose an Attention-Aggregating Feature Pyramid Network ($A^2$FPN) by combining channel-wise, spatial-wise and contextual attention modules (CSCA), which allows dynamic feature interaction and fusion.
\item With data collected from real-world surveillance cameras, we construct a Real-world Traffic Object Detection (RTOD) dataset. Incorporating more object variances and higher scenario complexities, it serves as a challenging benchmark to evaluate the model generalizability and scalability in real-world scenarios.
\item Extensive experimental evaluations on both public and real-world datasets demonstrate the superiority of MDDCNet to a wide range of mainstream detection models, which verifies the effectiveness and practicability of our method.
\end{highlights}


\maketitle

\section{Introduction}\label{sec1}

With the rapid development of intelligent transportation systems and autonomous driving technology, target detection in complex traffic scenes has become an important research direction in computer vision~\cite{TODYOLOv7IS25,TSDYOLO25SP}. However, accurate and efficient object detection in complex traffic scenes is still challenged by significant variances in diverse object classes including including vehicles, pedestrians, cyclists, and traffic signs irregularly distributed in scattered background.

Recent years have witnessed dramatic progress made in deep detection models which are typically represented by convolutional neural networks (CNNs)~\cite{RCNN,FastRCNN,FasterRCNN}. In particular, YOLO-series detectors have undergone continuous evolution since its introduction, achieving tremendous success due to their efficient end-to-end detection framework and multiscale feature fusion mechanism~\cite{TSDYOLO25SP}. Nevertheless, YOLO-like detection methods still suffer from limited long-range dependency modeling capability and fail to capture global contextual information, since they primarily rely on the local receptive fields of CNNs for feature modeling with inherent inductive biases. In complex traffic scenarios involving dense distributions of small objects in cluttered background, these methods are still prone to missed detections with false alarms, restricting further improvements in detection performance.

To compensate for the above-mentioned limitations, Vision Transformer (ViT) leverages the self-attention mechanism for effectively capturing global contextual information, exhibiting superior long-range modeling capability. \cite{DETR,DeformableDETR,ConditionalDETR,BHViT}. Additionally, hybrid architectures combining ViT with CNNs have been developed to strengthen model's global modeling. However, the self-attention mechanism in ViT incurs quadratic computational complexity, resulting in significant computational overhead for dense object detection tasks and posing great challenges to deployment of real-time traffic applications.

Recently, Mamba~\cite{Mamba,MambaVision,PlainMamba,EfficientVMamba} architecture based on the State Space Model (SSM) has emerged as a promising paradigm alternative to ViT. By employing a selective state space mechanism, it achieves long-sequence modeling with linear time complexity, maintaining high computational efficiency while demonstrating robust global modeling capabilities. In real-world traffic scenarios, fine-grained local structures are crucial for accurate localization of varying-scale objects with dense distributions. Therefore, it is insufficient to adequately capture detailed local information with sequential mechanism of Mamba, while pure convolutional operations struggle to effectively model global semantic relationships. Consequently, it is necessary to develop a collaborative framework that simultaneously accounts for both local detail modeling and global contextual representation in complex traffic scenes characterized by substantial multi-scale variations.

In this study, we have proposed a hybrid architecture for accurate multi-scale traffic detection termed as MDDCNet, which is short for Mamba with Deformable Dilated Convolution Network. Achieving CNN-Mamba synergy, our MDDCNet constructs a hierarchical hybrid backbone consisting of consecutive Multi-Scale Deformable Dilated Convolution (MSDDC) and Mamba blocks. In the high-resolution shallow stages, modeling local spatial structures with scale variations is strengthened with MSDDC blocks, whilst Mamba blocks based on a selective state space modeling module are embedded, allowing efficient modeling of long-range dependencies and global semantic information in the low-resolution deeper stages. Furthermore, a Channel-Enhanced Feed-Forward Network (CE-FFN) is designed to improve feature representation capability through the collaboration of local depthwise convolution and global channel attention. To promote the cross-scale feature fusion and interaction, an attention-enhanced Feature Pyramid Network based on Mamba is constructed to achieve dynamic fusion and optimization of multi-scale features comprehensively. These designs collectively contribute to improving the overall detection performance of the proposed model in complex traffic scenarios. To summarize, main contributions of this study are fivefold as follows:

\begin{itemize}
    \item In this study, we have proposed a novel detection framework termed MDDCNet tailored for traffic object detection. Within MDDCNet, a hierarchical hybrid backbone containing successive MDDCNet and Mamba blocks is well-designed, such that our framework can simultaneously capture local cues and model long-range dependencies.
    \item Within our MDDCNet and Mamba blocks, we have developed a channel-enhanced feedforward network (CE-FFN) to enable adaptive channel reweighting while preserving strong local spatial modeling. This allows our model to enjoy discriminative feature representation and hierarchical information interaction. 
    \item For detection neck, we have proposed an Attention-Aggregating Feature Pyramid Network ($A^2$FPN) based on Mamba. By combining channel-wise, spatial-wise and contextual attention modules (CSCA), our $A^2$FPN enables dynamic feature fusion, significantly benefiting accurate detection of varying-scale objects in complex traffic scenario.
    \item With data collected from real-world surveillance cameras, we construct a Real-world Traffic Object Detection (RTOD) dataset. Different from existing mainstream public datasets, RTOD incorporates more object variances and higher scenario complexities, thereby serving as a challenging benchmark to evaluate the model generalizability and scalability in real-world traffic scenarios.
   \item Extensive experimental evaluations on both public and real-world datasets demonstrate the superiority of MDDCNet to a wide range of mainstream detection models, which verifies the effectiveness and practicability of our method.
\end{itemize}

\section{Related work}\label{sec2}
\subsection{YOLO series algorithms}\label{subsec1}
Originated from~\cite{YOLOv1}, the YOLO-series detection models formulate object detection as a unified regression framework for prioritizing efficiency and end-to-end inference, facilitating practical detection tasks~\cite{FasterRCNN}. Generally comprising feature extraction backbone, multi-scale fusion neck, and detection head for prediction, the YOLO architecture has evolved rapidly by introducing numerous advanced designs, including batch normalization \cite{YOLOv2}, residual learning, multi-scale detection \cite{YOLOv3}, as well as the effective structures, e.g., CSPDarknet53 \cite{YOLOv4} and the lightweight focus module \cite{YOLOv5,YOLOv8}. However, in complex traffic scenarios, YOLO models suffer from insufficient adaptive feature weighting, and their inherent lightweight design impairs the perception of local fine-grained details. While recent state-space model integrations like MambaYOLO \cite{MambaYOLO} improve the model capability in global modeling, they still struggle to simultaneously capture fine local details and achieve cross-scale semantic alignment.

\subsection{Mamba-based detector}
Mamba \cite{Mamba} is a state-space model that captures long-range dependencies via hidden state evolution. Its continuous parameters are discretized via zero-order hold, allowing efficient recurrent or parallel global convolution formulations. Recent vision adaptations include Vision Mamba \cite{VisionMamba} with a bidirectional processing mechanism, VMamba \cite{VMamba} with a cross-scan strategy, and MambaVision \cite{MambaVision} that integrates self-attention in a hybrid architecture. In object detection, Mamba alleviates the limitations of CNNs in limited receptive fields and Transformers incurring quadratic complexity by combining linear complexity with effective long-range modeling. Since early 2024, Mamba-based detectors have rapidly evolved. ViM \cite{VisionMamba} introduced bidirectional SSM for 2D images. LocalMamba \cite{LocalMamba} performed SSM within local windows, outperforming Swin Transformer. VMamba \cite{VMamba} introduced Cross-Scan Module for spatial capture. EfficientVMamba \cite{EfficientVMamba} reduced cost via dilated convolution sampling. PlainMamba \cite{PlainMamba} demonstrated strong capacity with streamlined design. Beyond backbones, Mamba enables efficient detection heads, as exemplified by MambaYOLO, with extensions to video and remote sensing, e.g., RS-Mamba \cite{RSMamba}). In summary, Mamba-based detectors offer linear complexity and strong long-range modeling, showcasing significant potential for complex traffic scenarios requiring global contextual understanding.

\subsection{Attention-based feature interaction}
Channel attention methods, such as ECA \cite{ECANet} and MLCA \cite{MLCA}, enhance discriminative features by modeling inter-channel dependencies, while spatial attention mechanisms (e.g., the spatial attention module in CBAM \cite{CBAM}) focus on target regions and contextual information. Their integration enables comprehensive feature enhancement, demonstrating robust detection performance in complex scenarios. For multi-scale detection, such attention mechanisms are widely incorporated into feature pyramid networks to mitigate cross-scale semantic gaps, allowing dynamic feature recalibration that boosts the detection accuracy of small and scale-variant objects. However, mainstream attention-based feature interaction schemes, which usually built upon convolutions or self-attention, incur heavy computational overhead, making it difficult to balance global context modeling and efficiency. Therefore, achieving effective cross-scale feature interaction and contextual enrichment with low computation remains a critical challenge for object detection in complex traffic scenarios.


\section{Methodology}\label{sec3}
Densely distributed multi-scale objects in complex traffic scenes are prone to cluttered background, thereby exhibiting significant variances in appearances. To achieve accurate detection of multi-scale traffic objects, we propose a novel detector termed MDDCNet which is short for Mamba with Deformable Dilated Convolutions Network in this study, which will be detailed as follows.

\begin{figure*}
\centering
\includegraphics[width=\textwidth]{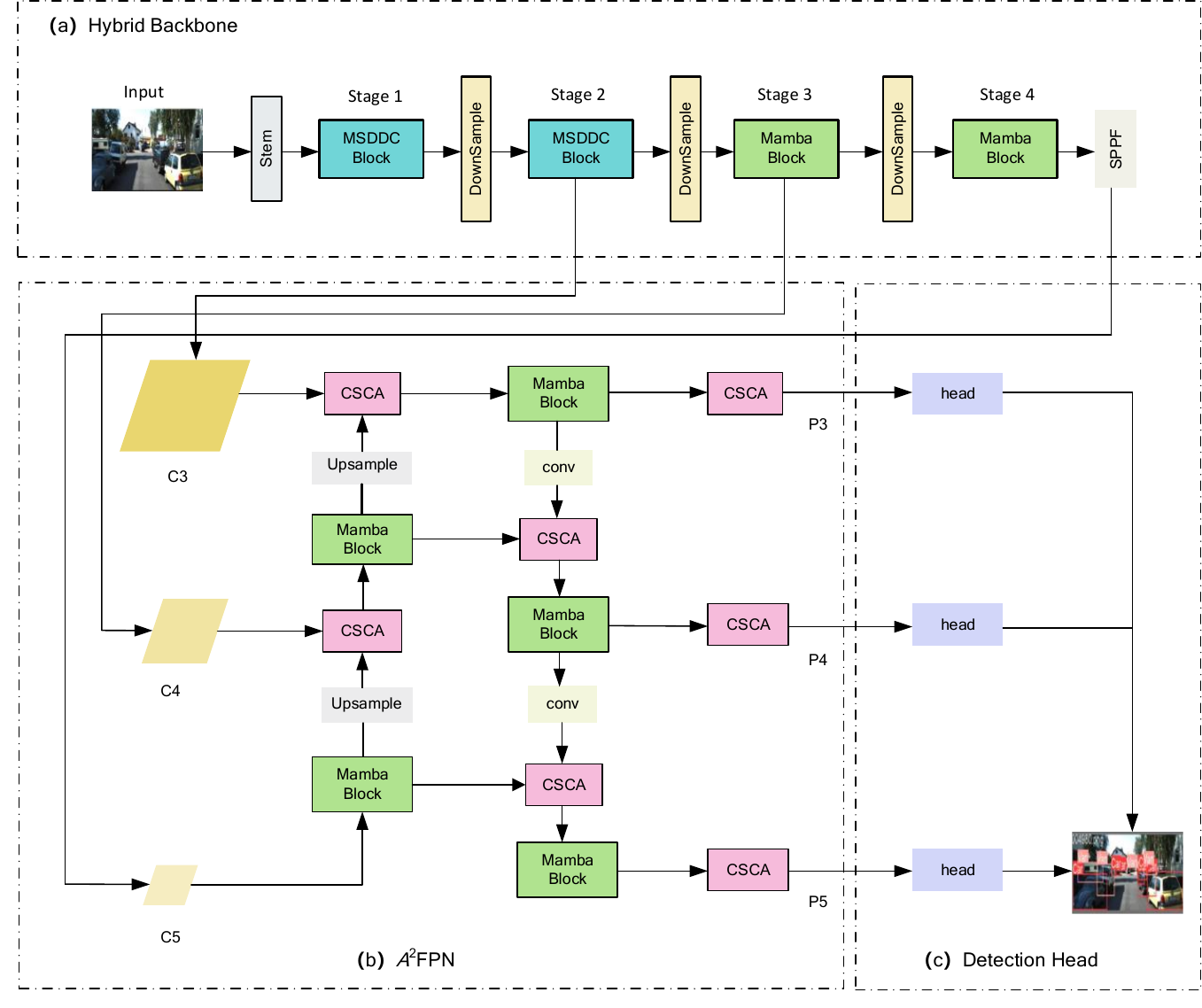}
\caption{The model architecture of our proposed MDDCNet which comprises Hybrid Backbone (a), Attention-Aggregating Feature Pyramid Network ($A^2$FPN) (b) and Detection Head (c).}
\label{fig:framework}
\end{figure*}

\subsection{Framework}
The overall framework of our MDDCNet is shown in Fig.~\ref{fig:framework}. Within MDDCNet, we devise a hierarchical backbone network with hybrid CNN-Mamba architecture from scratch. In backbone, a Multi-Scale Deformable Dilated Convolution module (MSDDC) is designed to enhance the local details and deformation modeling capability in the shallow high-resolution stage introduces, whilst the Mamba Block based on the selective state-space model is employed to efficiently capture the global contextual semantics with linear computational complexity the in deep low-resolution stage. With the resulting feature maps, an Attention-Aggregating Feature Pyramid Network ($A^2$FPN) based on Mamba Block is designed to dynamically enhance multi-scale feature interaction by fusing spatial-wise, channel-wise and contextual attention. Final detection head is imposed on the attention enhanced features, achieving end-to-end object detection. 

\begin{figure*}
\centering
\includegraphics[width=\textwidth]{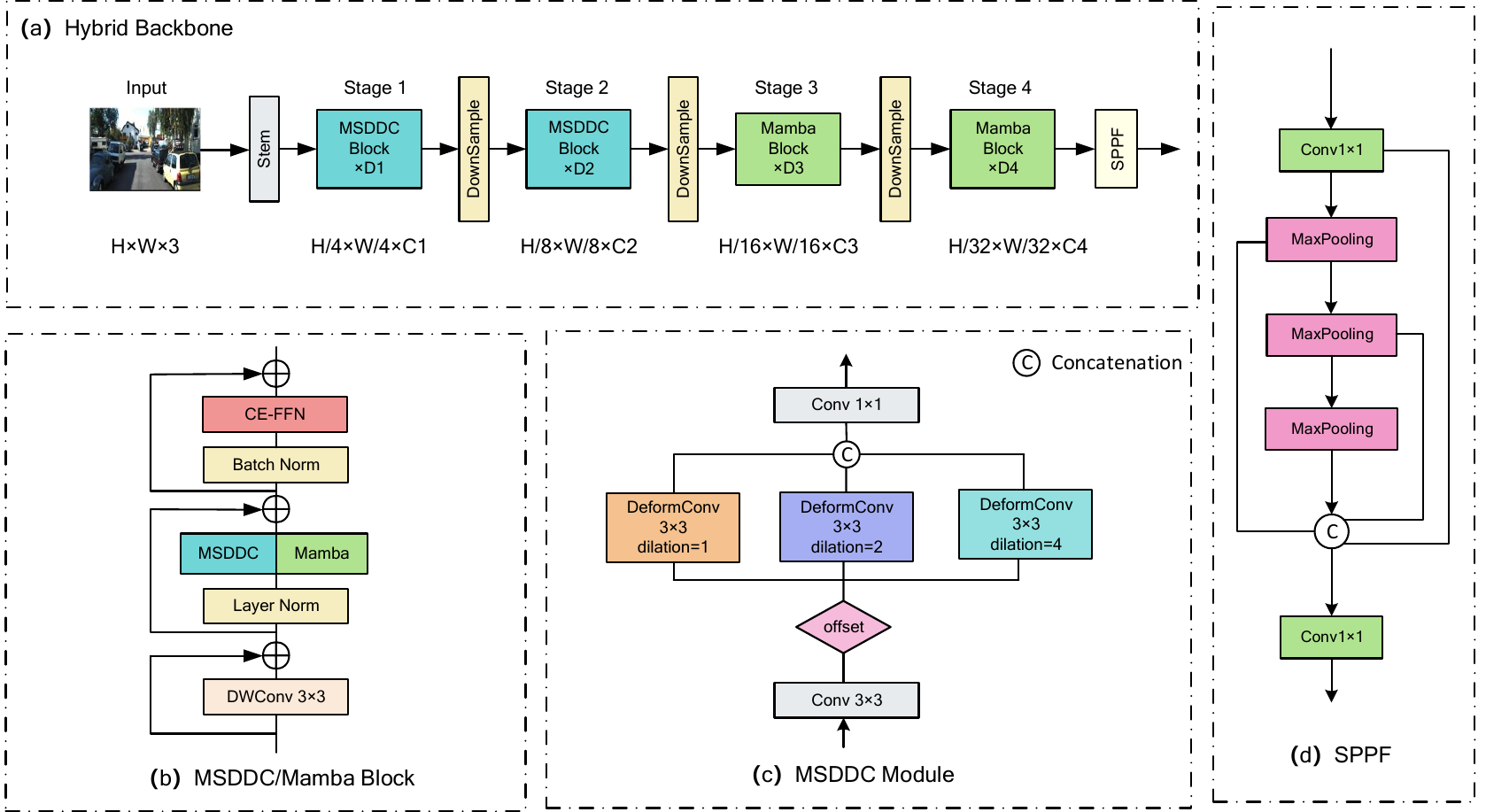} 
\caption{Illustration of the hybrid backbone network consisting of successive MSDDC and Mamba blocks.}
\label{fig:backbone}
\end{figure*}

\subsection{Hybrid Backbone Network}
To handle the challenge of multi-scale object distribution and contextual dependency in complex traffic scenarios, we propose a hierarchical CNN-Mamba hybrid backbone to simultaneously capture local details and model global semantics. As shown in Fig.~\ref{fig:backbone}, the backbone consists of successive Multi-Scale Deformable Dilated Convolutions (MSDDC) blocks in the shallow high-resolution stages and Mamba blocks in the deep low-resolution stages.

Given the input image \(X \in \mathbb{R}^{H \times W \times 3}\), it undergoes convolution embedding and batch normalization processing to generate the initial feature sequence, whilst a learnable position embedding is superimposed. The backbone network consists of four progressive downsampling stages (Stage 1-4), and in each stage, different feature processing strategies are executed:

\begin{equation}
H_l = 
\begin{cases}
\text{MSDDC-Block}(X_{l-1}) + X_{l-1}, & \text{Stage} \in [1,2] \\
\text{Mamba-Block}(X_{l-1}) + X_{l-1}, & \text{Stage} \in [3,4]
\end{cases}
\end{equation}

In the first place, we develop Multi-Scale Deformable Dilated Convolution (MSDDC) module to focus on local perception, since feature maps at high-resolution stages encode abundant spatial details. To be specific, multi-branch convolutions with deformable sampling within MSDDC can capture local texture and edge features, thereby providing fundamental geometric information for small object detection. In the latter low-resolution stages, feature responses encode larger receptive fields of the input image, thus carrying high-level semantic information. For global modeling, Mamba block is introduced to capture cross-region long-range dependencies through selective state-space matrices while maintaining linear sequence complexity, facilitating real-time target understanding in complex traffic scenarios.

\begin{table}[t]
    \centering
    \caption{Configurations of different model variants in our MDDCNet family.}
    \label{tab:MDDCNet_family}
    \resizebox{\textwidth}{!}{
    \begin{tabular}{c|c|c|c}
    \hline
       Model variants & MDDCNet-N & MDDCNet-T & MDDCNet-B\\
       \hline
       Embed. Dim. & [16, 32, 64, 128] & [32, 64, 128, 256] & [64, 128, 256, 512] \\
       \hline
       Stage 1 (H/4 $\times$ W/4) & MSDDC $\times$ 3 & MSDDC $\times$ 3 & MSDDC $\times$ 3 \\
       \hline
       Stage 2 (H/8 $\times$ W/8) & MSDDC $\times$ 3 & MSDDC $\times$ 3 & MSDDC $\times$ 3 \\
       \hline
       Stage 3 (H/16 $\times$ W/16) & Mamba $\times$ 9 & Mamba $\times$ 9 & Mamba $\times$ 12 \\
       \hline
       Stage 4 (H/32 $\times$ W/32) & Mamba $\times$ 3 & Mamba $\times$ 3 & Mamba $\times$ 3 \\
       \hline
       Params (M) & 4.8 & 6.6 & 18.0 \\
       \hline
       FLOPs (G) & 10.2 & 12.9 & 39.6 \\
       \hline       
    \end{tabular}
    }    
\end{table}

This cascaded backbone design provides a progressive feature learning pipeline from characterizing low-level local details to encoding high-level global semantics, thus allowing both accurate small-object detection and comprehensive complex scene understanding. In particular, we develop a family of MDDCNet models with varying model sizes to facilitate model scalability for practical applications. As presented in Table~\ref{tab:MDDCNet_family}, the scale of our MDDCNet grows with gradually increasing channel dimensions and cascaded blocks, yielding three variants, namely the lightweight MDDCNet-N, medium-sized MDDCNet-T and larger MDDCNet-B. 

\subsubsection{MSDDC Module}
In MSDDC block, we design a Multi-Scale Deformable Dilated Convolution Module (MSDDC) as shown in Fig.~\ref{fig:backbone}(c), allowing the shallow layers of the network to perceive local details and multi-scale targets in traffic scenes. Compared to traditional CNNs, simple multi-scale convolutions or deformable convolutions, MSDDC achieves the synergy of both multi-scale receptive fields and deformable adaptive sampling, facilitating object detection in complex traffic scenes. 

Given the input feature \(X_{l-1}\), the processing pipeline of MSDDC module includes three steps as follows.

\paragraph{Offset Generation}
In order to realize adaptive sampling of deformable convolution, a spatial offset $\Delta P$ is obtained by $3 \times 3$ convolution as:
\begin{equation}
\Delta P = \text{Conv}_{3 \times 3}(X_{t-1})
\end{equation}
where $\Delta P \in \mathbb{R}^{H \times W \times 18}$  to the two-dimensional offsets of all the sampling points of each $3 \times 3$ convolution kernel. Compared to the fixed sampling of traditional CNNs, MSDDC can adaptively adjust the convolution kernel positions according to the object shape, thus improving the robustness to vehicle appearance changes, partial occlusions, and non-rigid objects.

\paragraph{Parallel Multi-scale Deformable Dilated Convolutions}
In order to capture cross-scale features, three parallel branches of deformable convolutions with dilation rates $d \in \{1, 2, 4\}$ are designed, each using the same offset $\Delta P$:
\begin{equation}
F_d = \text{DeformConv}_{3 \times 3}^d (X_{i-1}, \Delta P), \quad d \in \{1, 2, 4\}
\end{equation}

On the one hand, traditional multi-scale convolutions only obtain different receptive fields through different convolution kernels or dilation rates, whereas the sampling location is fixed and insensitive to object shape changes. On the other hand, simple deformable convolutions can achieve adaptive sampling, whereas they usually operate only at a single scale, which makes it difficult to simultaneously take into account the information of both close and distant objects. In contrast, our MSDDC integrate both multi-scale convolution and deformable convolution to comprehensively capture important cues ranging from local details of the near vehicle to the overall contour of the distant object, thereby providing rich basic features for subsequent detection.

\paragraph{Multi-scale Fusion}
The outputs of the three branches are concatenated along the channel dimension, then fused and downsampled via a $1 \times 1$ convolution to produce the final features:
\begin{equation}
F_{\text{MSDDC}} = \text{Conv}_{1 \times 1} \left( \text{Concat}(F_1, F_2, F_4) \right)
\end{equation}


\begin{figure}
\centering
\includegraphics[width=\textwidth]{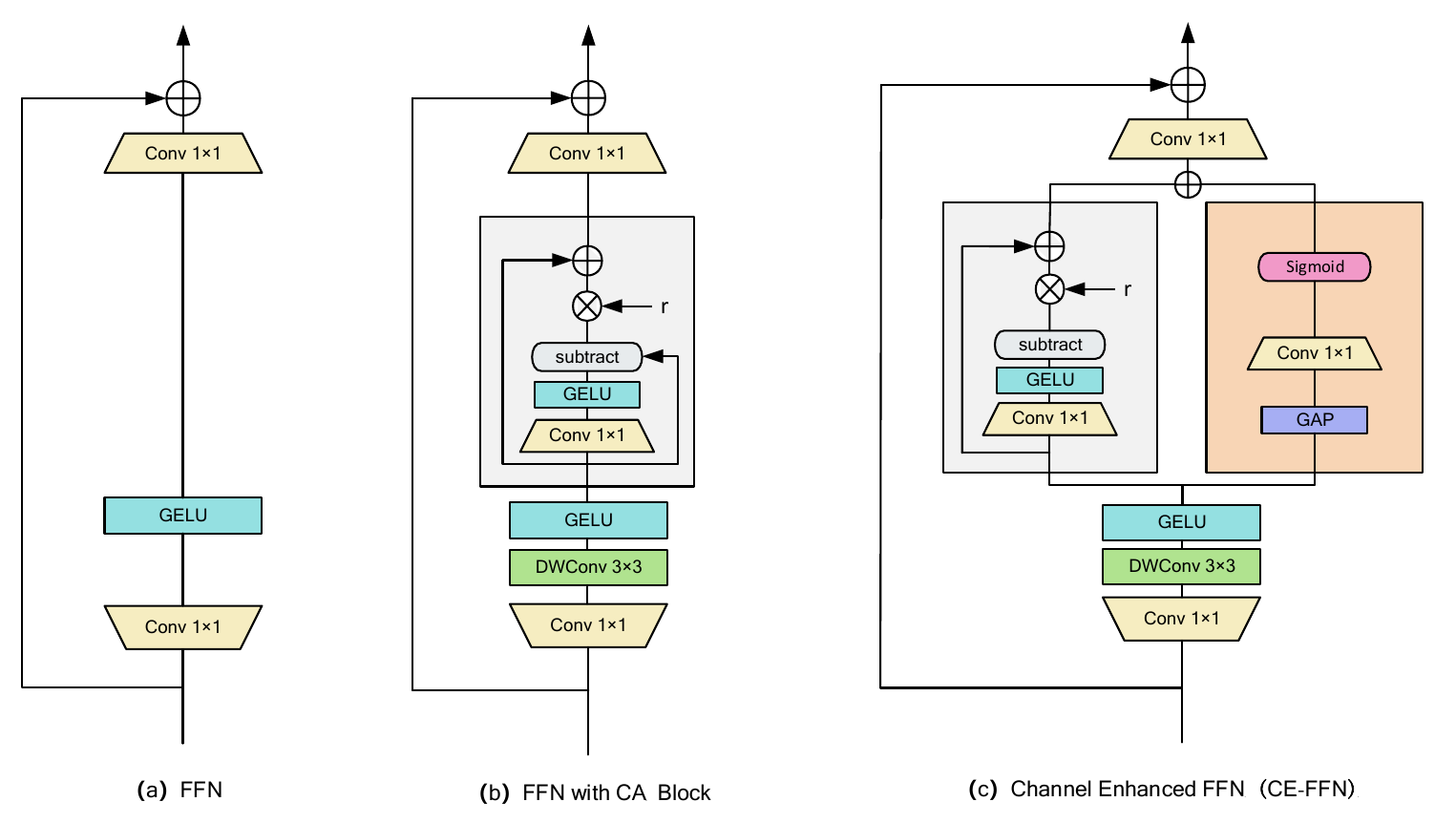} 
\caption{Comparison of our proposed CE-FFN (c) and the vanilla FFN (a) and FFN with CA-Block (b).}
\label{fig:comp_FFN}
\end{figure}

\subsubsection{Mamba Module}
In latter stages in our backbone network, Mamba module based on the selective SSM is introduced to replace the traditional Transformer architecture with self-attention mechanism, such as efficient long sequence modeling can be achieved. With the help of the selective scanning mechanism, Mamba is able to efficiently capture global long-range dependencies, such as the road topology, relative vehicle positions, and motion relationships in the traffic scenarios, while maintaining linear computational complexity. More importantly, it is complementary to the shallow MSDDC module, allowing our network to achieve both fine-grained local perception and global semantic understanding in complex traffic scenarios, thus significantly boosting the detection performance.


\begin{figure}
\centering
\includegraphics[width=\textwidth]{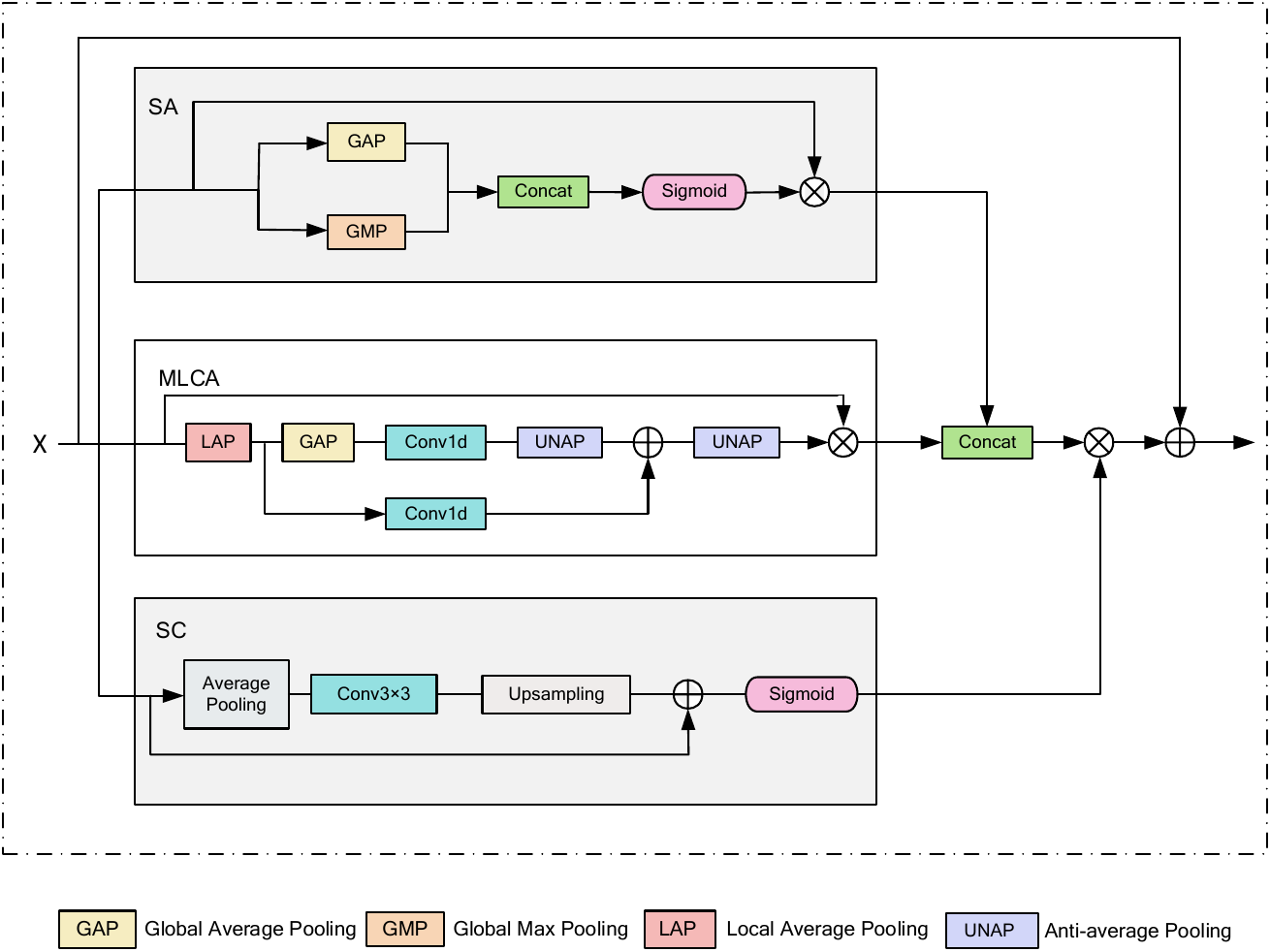} 
\caption{Our proposed Contextual-Spatial-Channel Attention (CSCA) synergy module. Comprising three complementary attention branches, it leverages an attention-aggregating mechanism for comprehensively strengthening feature discriminative power, which significantly benefits multi-scale feature fusion and interaction.}
\label{fig:CSCA}
\end{figure}

\subsubsection{CE-FFN}
Traditional Feedforward Network (FFN) in ViT lacks explicit local spatial modeling, making it difficult to characterize fine-grained details such as vehicle edges and contours in complex traffic scenarios. Moreover, its vulnerability to object scale variances and cluttered backgrounds further impairs its feature representation ability. Taking advantage of Channel Aggregation (CA) block, we design a Channel-Enhanced Feedforward Network (CE-FFN) as illustrated in Fig.~\ref{fig:comp_FFN}(c). The proposed CE-FFN incorporates a global branch complementary to the local branch based on the CA block to achieve local-global collaborative modeling, which significantly enhances the feature representation capacity.

Given the input feature $X \in \mathbb{R}^{C \times H \times W}$, the proposed CE-FFN is formulated as:

\begin{equation}
\begin{aligned}
  Y &= \text{GELU}\left({\text{DWConv}}_{3 \times 3}\left(\text{Conv}_{1 \times 1}(X)\right)\right) \\
F_{\text{CE\_FFN}} &= \text{Conv}_{1 \times 1} \left( F_{\text{global}}(Y) + F_{\text{local}}(Y) \right) + X  
\end{aligned}
\end{equation}

Inheriting the idea of local aggregation from CA Block, the local branch is formulated as:

\begin{equation}
F_{\text{local}} = r \odot \left( Y - \text{GELU}(\operatorname{Conv}_{1 \times 1}(Y)) \right) + Y
\end{equation}

Through channel transformation with a residual modulation mechanism, the local branch enhances the local structure and detailed features of the object-aware region while suppressing redundant background noise.

Moreover, the devised global branch obtains global semantic embedding via global average pooling and generates channel-wise weights to guide feature recalibration, which is formulated as:

\begin{equation}
F_{\text{global}} = \sigma\left(\text{Conv}_{1 \times 1}\left(\text{GAP}(Y)\right)\right)
\end{equation}
where $\sigma(\cdot)$ denotes the Sigmoid activation function and GAP refer to the global average pooling operation. As a complement to the local branch based on the CA block, the global branch can effectively model global contextual information, thereby enhancing the model ability to capture scale variations and global semantic relationships.

Fig.~\ref{fig:comp_FFN} intuitively compares three different architectures of vanilla FFN, FFN with CA block and our CE-FFN. It is shown that the proposed CE-FFN preserves the channel modeling capability of the vanilla FFN and local perception of the CA block, whilst strengthening the global semantic awareness, thus alleviating the limitation of background interference caused by relying only on local cues. This advantage of CE-FFN is manifested in simultaneously characterizing the structural details of nearby large vehicles and enhancing the semantic perception of distant small objects. By achieving local-global synergy, it further suppresses irrelevant interference in complex and cluttered background conditions. 


\subsection{Attention-Aggregating Feature Pyramid Network (A$^2$FPN)}
To maintain both high-level semantics and low-level cues in multi-scale traffic object detection, we construct a Mamba-based Attention-Aggregating Feature Pyramid Network ($A^2$FPN) as the detection neck by embedding Mamba blocks into the conventional FPN architecture, thus facilitating global context enhancement and long-range dependencies modeling across varying-scale features. Moreover, we develop a Contextual-Spatial-Channel Attention (CSCA) synergy module via an attention-aggregating mechanism. As illustrated in Fig.~\ref{fig:CSCA}, our CSCA module integrates three complementary attention branches, namely Spatial Attention (SA) branch, Mixed Local Channel Attention (MLCA) branch, and scale Self-Calibration (SC) branch, into the cross-scale FPN fusion and layer-wise output features, achieving adaptive modulation of spatial, channel and scale information. 

Within CSCA module, the first SA branch models discriminative spatial regions in the feature maps, which allows the network to focus on the key structural locations of traffic objects, e.g., vehicles and pedestrians. It further enhances object localization accuracy while suppressing interference from complex backgrounds. The second MLCA branch extends traditional channel modeling based on global average pooling by incorporating local spatial awareness. This enables the network to capture both global semantic dependencies and local inter-channel response variations. Compared to traditional channel attention mechanisms such as CA \cite{ECANet} that only rely on global statistical information, MLCA \cite{MLCA} can more effectively retain the semantic channels related to small objects and local details, avoiding excessive smoothing of fine-grained semantics during multi-scale fusion, thereby improving the accuracy and robustness of channel attention. The third SC branch introduces multi-scale contextual information to dynamically modulate the fused features, enabling the network to adaptively calibrate its semantic distribution during cross-scale feature interaction, thereby enhancing cross-scale consistency and overall perception ability.

The three complementary attention are aggregated, providing comprehensive feature enhancement and collaboratively benefiting multi-scale feature interaction. Given the input feature \(X \in \mathbb{R}^{C \times H \times W}\), CSCA adopts a residual structure to fuse the attention-enhanced features as follows:

\begin{equation}
Y = \text{Conv}_{1 \times 1} (\text{Concat}(\text{SA}(X), \text{MLCA}(X))) \odot \text{SC}(X) + X 
\end{equation}
where \(\odot\) denotes element-wise multiplication, while SA($\cdot$), MLCA($\cdot$), and SC($\cdot$) respectively represent the output of three attention branches.


\begin{figure}
    \centering  
    \begin{subfigure}[b]{0.48\textwidth}
        \centering
        \includegraphics[width=\linewidth]{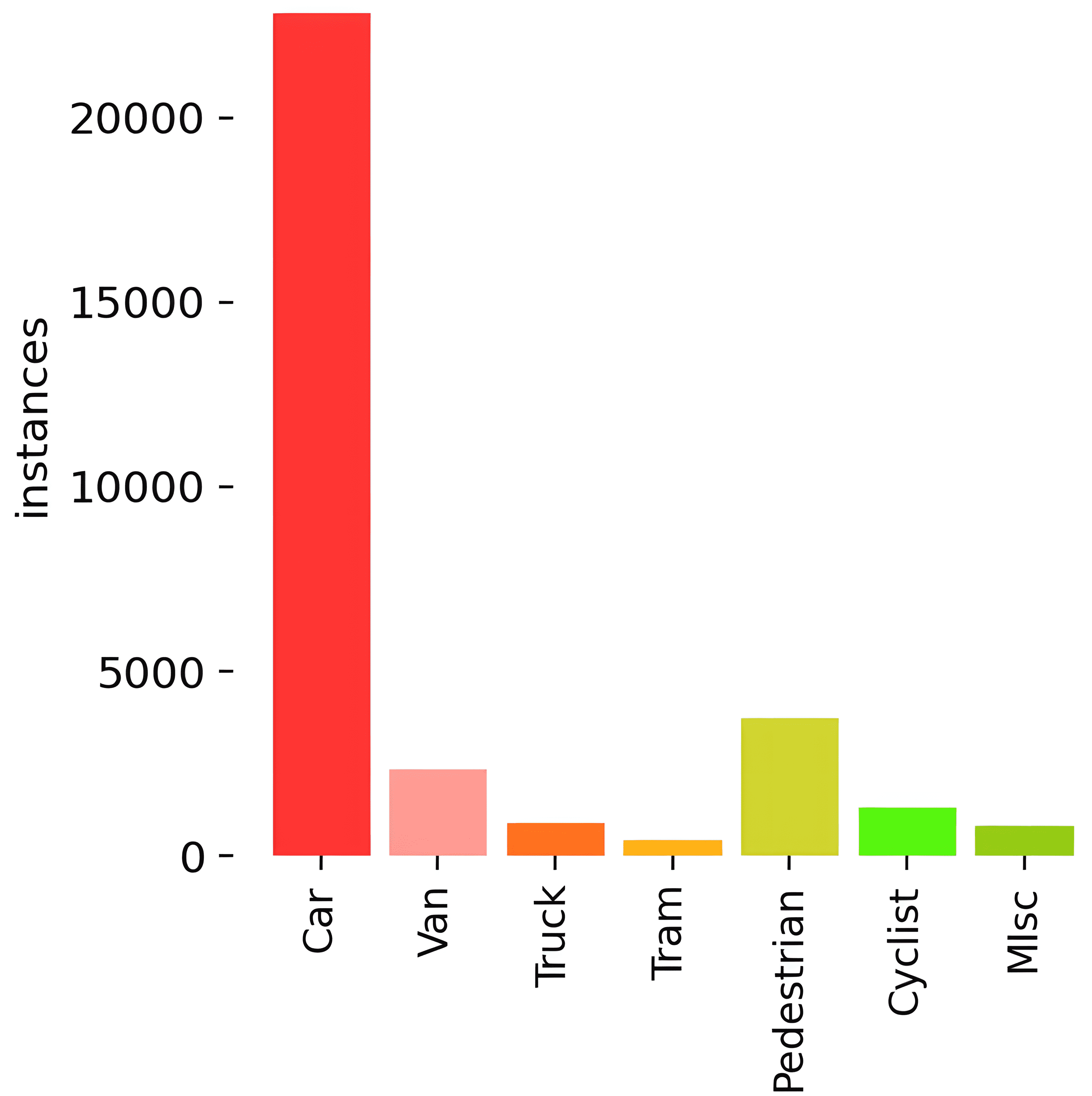}  
        \caption{}  
        \label{subfig5a}  
    \end{subfigure}
    \hfill  
    \begin{subfigure}[b]{0.48\textwidth}
        \centering
        \includegraphics[width=\linewidth]{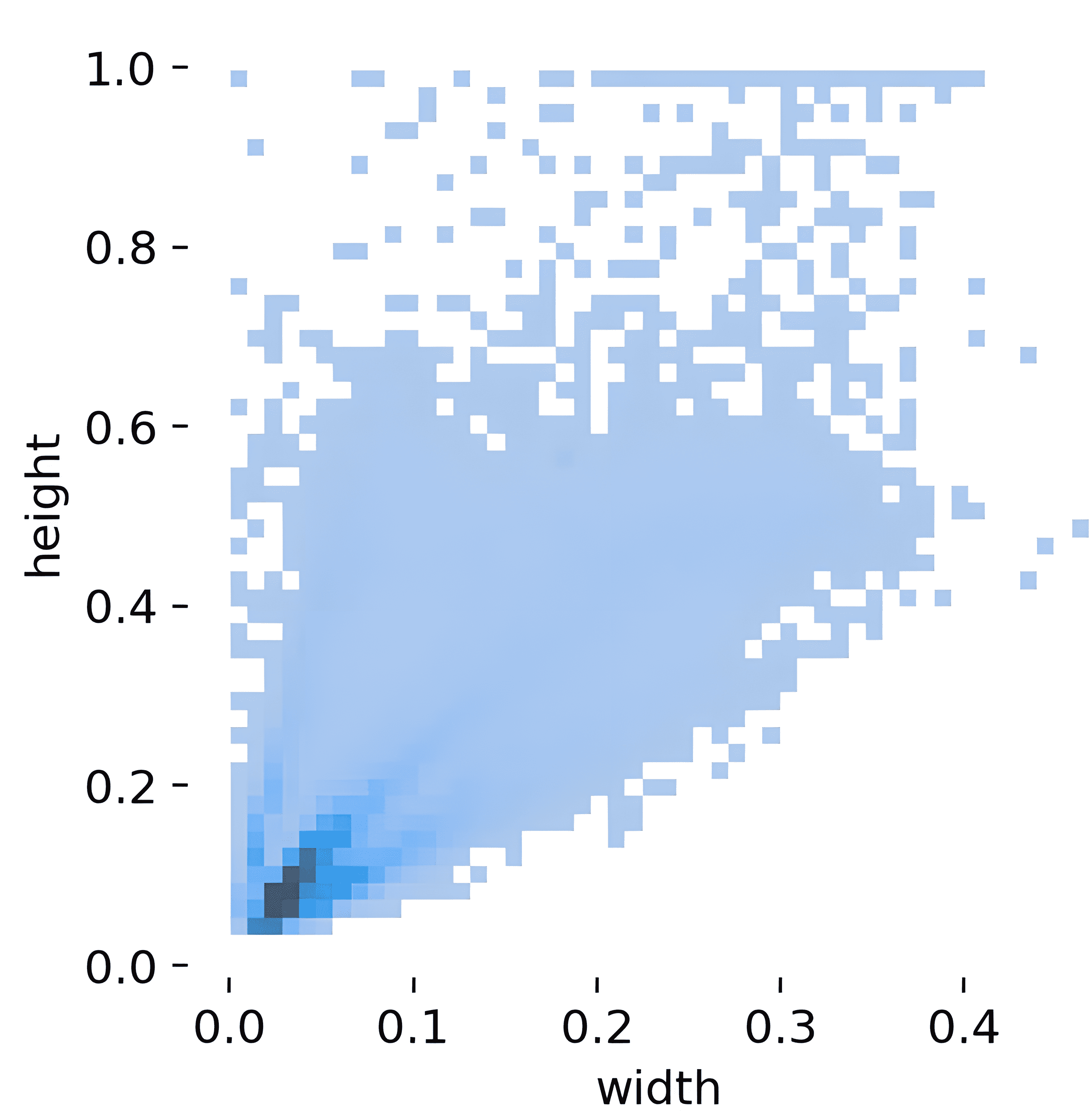}
        \caption{}
        \label{subfig5b}
    \end{subfigure}   
    \caption{Statistics of KITTI dataset in traffic object categories (a) and scale distributions (b).}
    \label{fig:KITTI_sta}  
\end{figure}

\begin{figure}
    \centering  
    \begin{subfigure}[b]{0.48\textwidth}
        \centering
        \includegraphics[width=\linewidth]{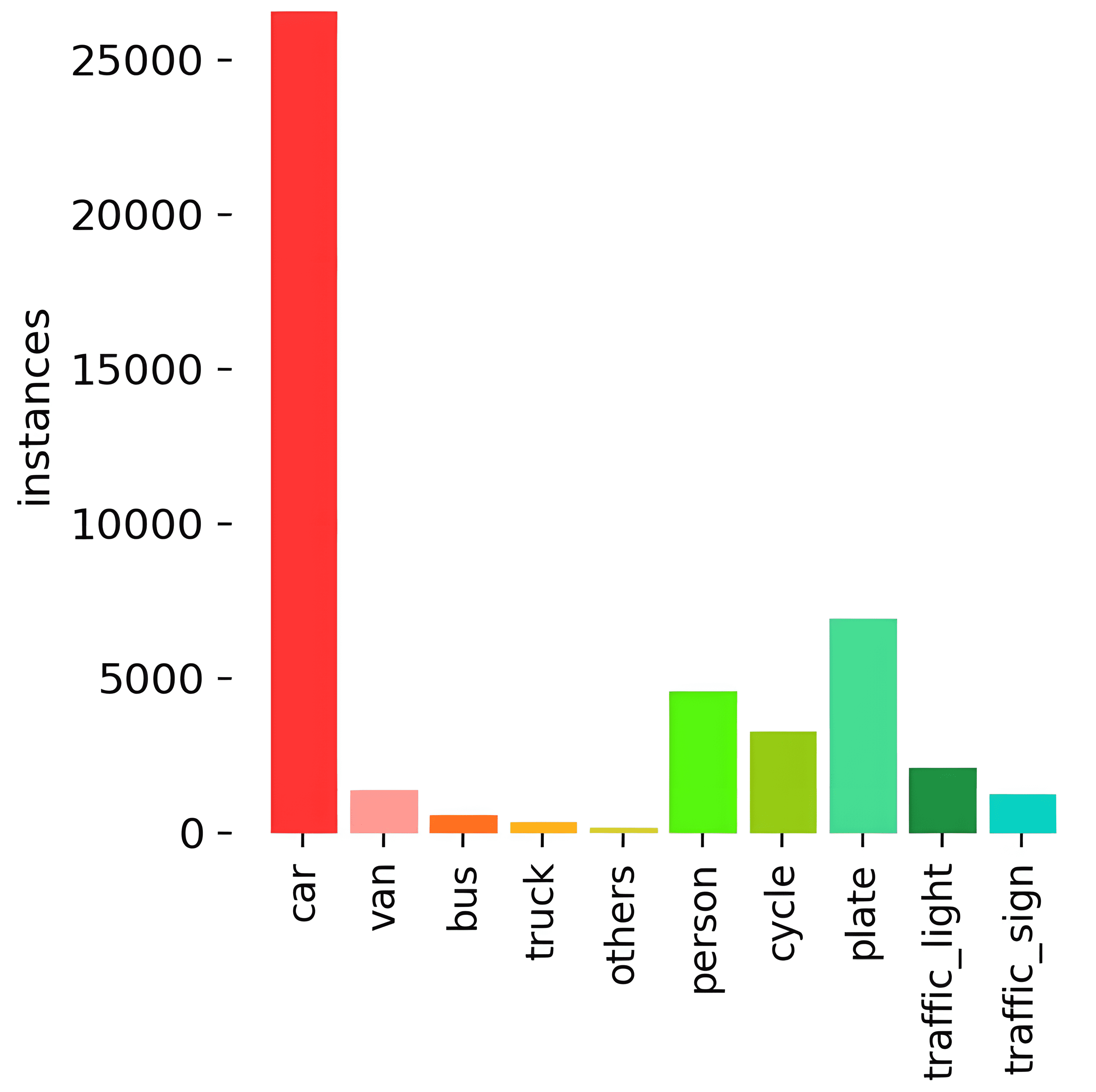}  
        \caption{}  
        \label{subfig6a}  
    \end{subfigure}
    \hfill  
    \begin{subfigure}[b]{0.48\textwidth}
        \centering
        \includegraphics[width=\linewidth]{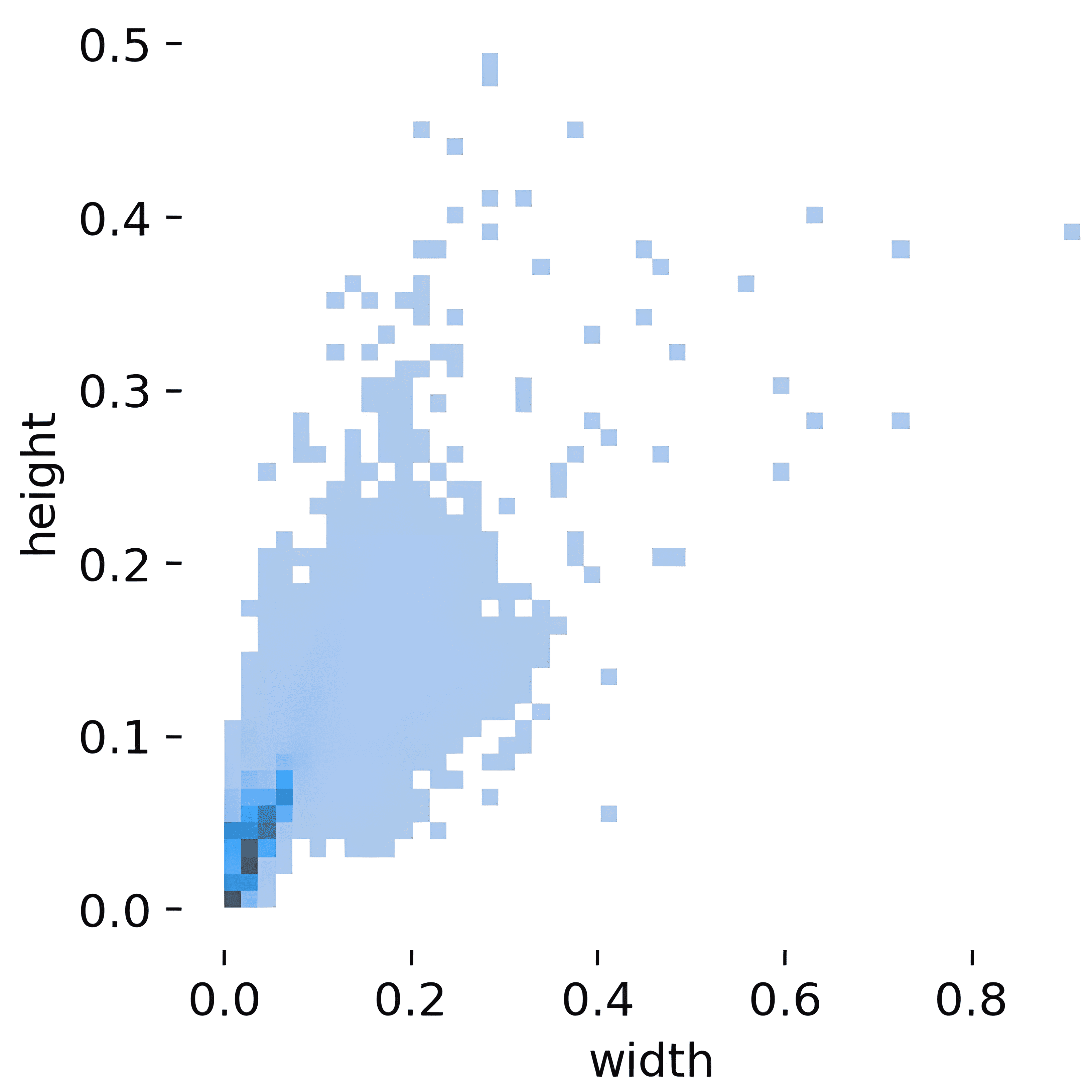}
        \caption{}
        \label{subfig6b}
    \end{subfigure}   
    \caption{Statistics of our RTOD dataset in traffic object categories (a) and scale distributions (b).}
    \label{fig6}  
\end{figure}

\begin{figure}
\centering
\includegraphics[width=\textwidth]{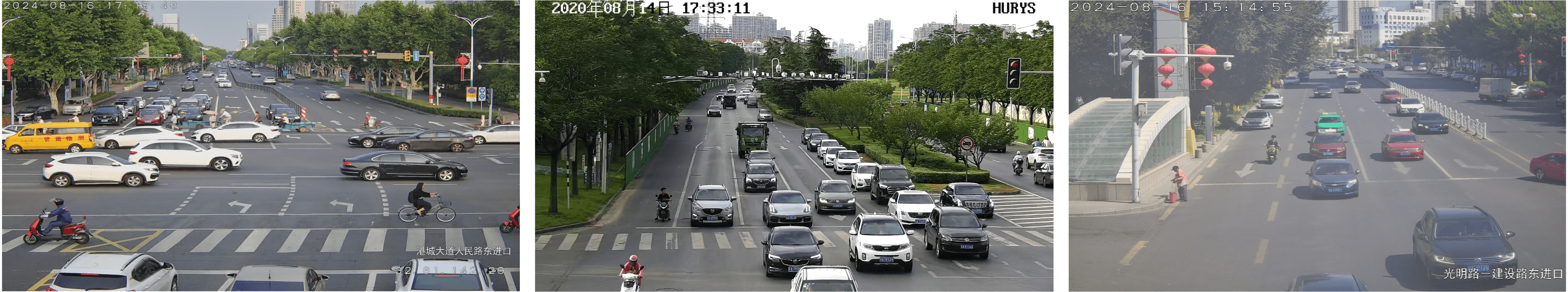} 
\caption{Example images in RTOD dataset.}
\label{fig:RTOD_exam}
\end{figure}

\section{Experiments}

\subsection{Datasets}
In our experiments, we have carried out extensive experimental evaluations on both public KITTI dataset and real-world RTOD dataset. As one of the most widely used public benchmarks in autonomous driving and computer vision, KITTI covers a wide range of common traffic object categories, e.g., car, van, pedestrian, which exhibit significant scale variances in spatial distribution as shown in Fig.~\ref{fig:KITTI_sta}. 

To further evaluate our method in practical applications, we construct and annotate a new dataset named RTOD (Real-world Traffic Object Detection). The dataset consists of 1,180 high-quality images sampled from real-time traffic surveillance videos captured on urban streets in a large city, reflecting typical real-world deployment scenarios. Compared to existing public datasets, our RTOD dataset further expands the object categories in traffic scenes to ten typical classes including car, van, bus, truck, person, cycle, plate, traffic light, traffic sign and others, thereby providing a more diverse distribution of current urban traffic participants as shown in Fig.~\ref{subfig6a}. Moreover, the images in RTOD pose greater challenges to accurate detection, since they are directly collected from real-world scenes, thus showcasing more dramatic variances in illumination, weather and viewpoint. Due to the the elevated camera positions and wide field of view, most objects appear at small scales as illustrated in Fig.~\ref{subfig6b}, making it more difficult to distinguish them from cluttered background. Therefore, it is well-suited for assessing the model practicability and generalization ability in real-world scenarios. Fig.~\ref{fig:RTOD_exam} illustrates representative images in RTOD dataset.

\subsection{Experimental Setup}
Both datasets are randomly partitioned into training, validation, and test sets with a ratio of 8:1:1 for respective model training, hyperparameter tuning, and performance evaluation. Moreover, both accuracy and efficiency metrics are involved for performance measure, including precision score, recall rate, mAP, FLOPs and FPS (Frame Per Second). For data preprocessing, all the input images are uniformly resized to 640 $\times$ 640 pixels to ensure consistent input dimensions. In the training stage, the Mosaic data augmentation strategy is adopted to enhance the model’s generalization ability and robustness, particularly in scenarios involving dense and small objects. During the training process, the momentum is set to 0.937, the initial learning rate to 0.01, while the batch size to 32. Additionally, stochastic gradient descent (SGD) is employed for the optimizer. To ensure adequate convergence, the model is trained for 150 epochs on the KITTI dataset and 300 epochs on the RTOD dataset. In our comparative studies, the competing methods include various mainstream detectors based on CNN and Mamba architectures, such as the YOLO-series detection models~\cite{YOLOv3,YOLOv5,YOLOv8,YOLOv9,YOLOv11,YOLOv12,YOLOv13}, and MambaYOLO~\cite{MambaYOLO}. In implementation, all the experiments are conducted on a desktop with a RTX 3090 GPU using PyTorch framework.

\begin{table*}
  \centering
  \caption{Comprehensive comparison on the KITTI Dataset. The results of our method are highlighted in bold. }
  \label{Tab:comprehensive_KITTI}
  \resizebox{\linewidth}{!}{
  \begin{tabular}{l|c|c|c|c|c|c|c}
    \hline
    \textbf{Detectors} & \textbf{P(\%)} & \textbf{R(\%)} & \textbf{mAP@50(\%)} & \textbf{mAP@50-95(\%)} & \textbf{Params(M)} & \textbf{FLOPs(G)} & FPS  \\
    \hline
    YOLOv3t & 87.7 & 75.3 & 82.1 & 58.3 & 12.1 & 18.9 & 67.52 \\
    \hline
    YOLOv5n & 91.3 & 77.8 & 88.5 & 63.3 & 2.5 & 7.2 & 80.64 \\
    \hline
    YOLOv6s & 89.8 & 76.4 & 85.3 & 60.6 & 4.2 & 11.9 & 80.65 \\
    \hline
    YOLOv8n & 90.5 & 80.9 & 89.5 & 65.1 & 3.0 & 8.2 & 85.32 \\
    \hline
    YOLOv8s & 93.1 & 87.6 & 93.3 & 73.0 & 11.1 & 28.5 & 66.67 \\
    \hline
    YOLOv10n & 83.7 & 79.1 & 85.9 & 62.5 & 2.7 & 8.4 & 90.50 \\
    \hline
    YOLOv10s & 91.2 & 85.6 & 91.9 & 71.2 & 8.0 & 24.5 & 70.66 \\
    \hline
    YOLOv11n & 88.2 & 79.6 & 88.4 & 74.9 & 2.6 & 6.3 & 90.90 \\
    \hline
    YOLOv12n & 90.6 & 77.4 & 86.9 & 62.0 & 2.5 & 5.8 & 89.28 \\
    \hline
    YOLOv13n & 88.4 & 81.4 & 88.7 & 65.9 & 2.4 & 6.2 & 87.72 \\
    \hline
    MambaYOLO-T & 92.2 & 84.7 & 91.6 & 69.9 & 5.9 & 13.6 & 73.76 \\
    \hline
    \hline
    \textbf{MDDCNet-N} & \textbf{90.3} & \textbf{84.7} & \textbf{92.0} & \textbf{68.3} & \textbf{4.8} & \textbf{10.2} & \textbf{76.33} \\ 
    \textbf{MDDCNet-T} & \textbf{95.3} & \textbf{86.1} & \textbf{93.3} & \textbf{72.3} & \textbf{6.6} & \textbf{12.9} & \textbf{73.02} \\ 
    \textbf{MDDCNet-B} & \textbf{95.3} & \textbf{89.2} & \textbf{94.1} & \textbf{74.3} & \textbf{18.0} & \textbf{39.6} & \textbf{50.65} \\
    \hline
  \end{tabular}
  }
\end{table*}

\subsection{Main results}
Table~\ref{Tab:comprehensive_KITTI} presents the results of our proposed MDDCNet and a comparison of different mainstream detectors on the KITTI dataset. It can be observed that MDDCNet achieves consistent superiority to other competitors. In particular, even our lightweight MDDCNet-N model outperforms MambaYOLO-T by 0.4\% mAP@50 with fewer parameters and faster inference speed. Moreover, MDDCNet-T reports 93.3\% mAP@50 with only 6.6M parameters and 12.9G FLOPs, surpassing MambaYOLO-T by 1.7\% with comparable computational overhead. This advantage also holds when compared to YOLO detectors, showcasing the strong competitiveness of our MDDCNet models against state-of-the-art YOLO variants. Our larger variant MDDCNet-B obtains the highest 94.1\% mAP@50 while maintaining favorable inference efficiency. Notably, YOLOv8s also achieves 93.3\% mAP@50, yet with almost 2$\times$ computational costs of our MDDCNet-T.


\begin{table*}
  \centering
  \caption{Comparison of different detectors across various categories on the KITTI dataset (mAP@50\%).}
  \label{Tab:class_KITTI}
  \resizebox{\linewidth}{!}{
  \begin{tabular}{l|c|c|c|c|c|c|c}
    \hline
    \textbf{Detectors} & \textbf{Car} & \textbf{Van} & \textbf{Truck} & \textbf{Tram} & \textbf{Pedestrian} & \textbf{Cyclist} & \textbf{Misc} \\
    \hline
    YOLOv3t & 88.7 & 86.9 & 97.3 & 87.8 & 68.8 & 65.8 & 79.7 \\
    \hline
    YOLOv5n & 95.2 & 91.5 & 96.9 & 92.2 & 76.5 & 80.7 & 86.4 \\
    \hline
    YOLOv6s & 94.9 & 87.9 & 95.6 & 91.0 & 76.7 & 73.9 & 77.1 \\
    \hline
    YOLOv8n & 95.7 & 93.1 & 95.8 & 92.0 & 79.3 & 78.6 & 92.0 \\
    \hline
    YOLOv8s & 97.3 & 96.8 & 98.8 & 94.3 & 84.8 & 85.8 & 95.2 \\
    \hline
    YOLOv10n & 94.7 & 90.3 & 96.2 & 91.0 & 73.0 & 77.0 & 79.2 \\
    \hline
    YOLOv10s & 96.7 & 95.7 & 98.2 & 94.5 & 83.3 & 84.0 & 91.2 \\
    \hline
    YOLOv11n & 95.5 & 92.1 & 96.4 & 91.1 & 77.0 & 79.4 & 87.6 \\
    \hline
    YOLOv12n & 94.9 & 90.2 & 94.6 & 94.3 & 75.2 & 78.8 & 80.6 \\
    \hline
    YOLOv13n & 95.4 & 91.2 & 97.4 & 90.8 & 79.2 & 80.1 & 86.7 \\
    \hline
    MambaYOLO-T & 96.9 & 94.9 & 97.0 & 94.2 & 83.7 & 83.4 & 91.0 \\
    \hline
    \hline
    \textbf{MDDCNet-N} & \textbf{96.8} & \textbf{94.7} & \textbf{99.1} & \textbf{96.1} & \textbf{82.5} & \textbf{81.1} & \textbf{93.8} \\
    \textbf{MDDCNet-T} & \textbf{97.6} & \textbf{96.4} & \textbf{99.1} & \textbf{94.4} & \textbf{86.6} & \textbf{85.2} & \textbf{94.0} \\
    \textbf{MDDCNet-B} & \textbf{97.5} & \textbf{96.8} & \textbf{98.4} & \textbf{98.2} & \textbf{86.5} & \textbf{87.1} & \textbf{94.2} \\
    \hline
  \end{tabular}
  }
\end{table*}

Table~\ref{Tab:class_KITTI} provides a detailed comparison of the detection accuracy across different categories in the KITTI dataset for various models. It can be seen that our varying-scale MDDCNet models unanimously exceed 94\% mAP@50 when detecting vehicle-related objects (car, van, truck, tram), indicating that the proposed method has stronger modeling capabilities and discriminative capabilities. For small-object categories such as pedestrians and cyclists, our MDDCNet-T model outperforms YOLOv10s by approximately 3\% and 1\% respectively, effectively mitigating the issue of small-target missed detection. This validates the model's enhanced ability to capture fine-grained features and contextual information.

\begin{figure*} 
  \centering
  \begin{subfigure}{0.48\textwidth}
    \centering
    \includegraphics[width=\textwidth]{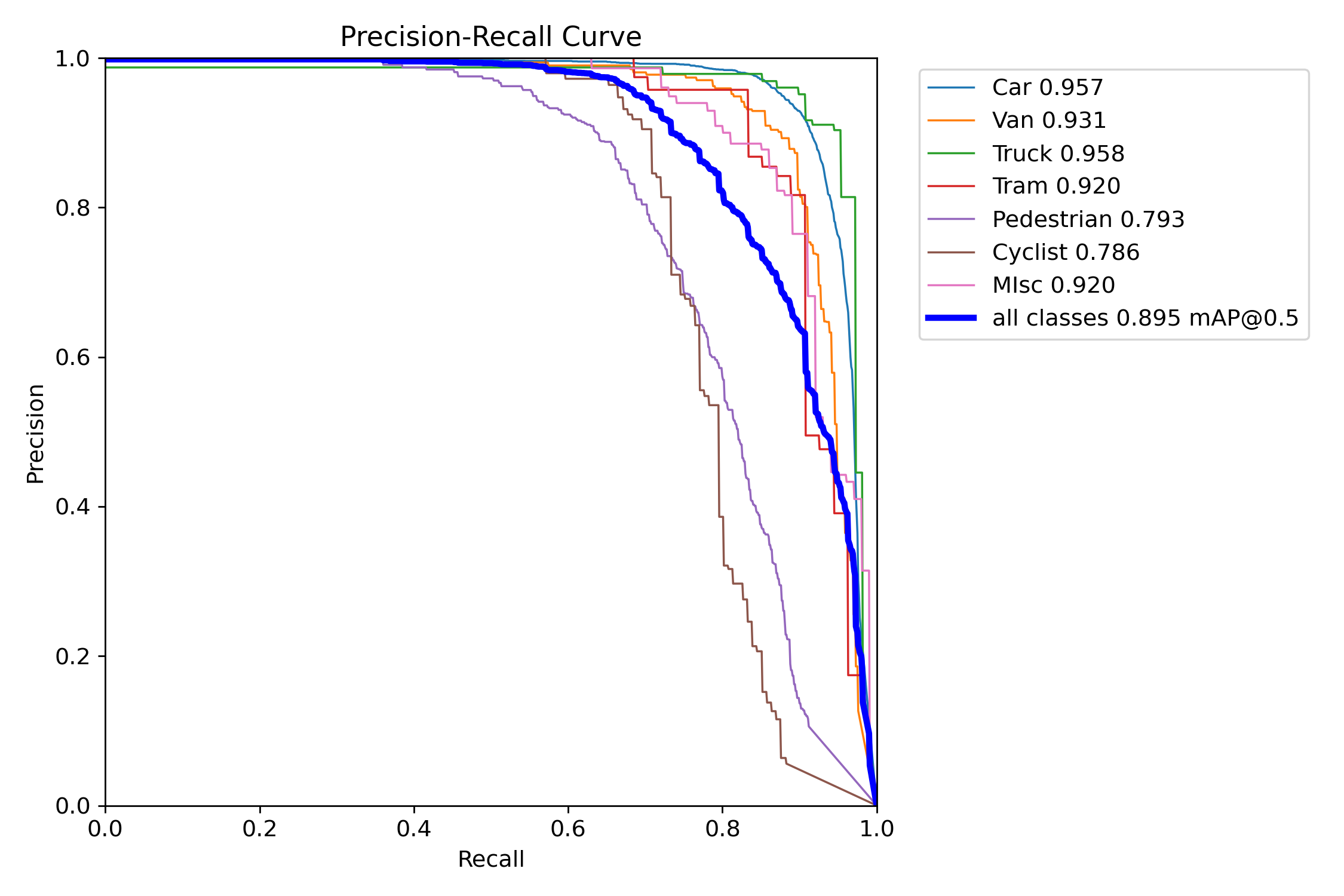}
    \caption{}  
    \label{fig8-1}
  \end{subfigure}
  \hfill
  \begin{subfigure}{0.48\textwidth}
    \centering
    \includegraphics[width=\textwidth]{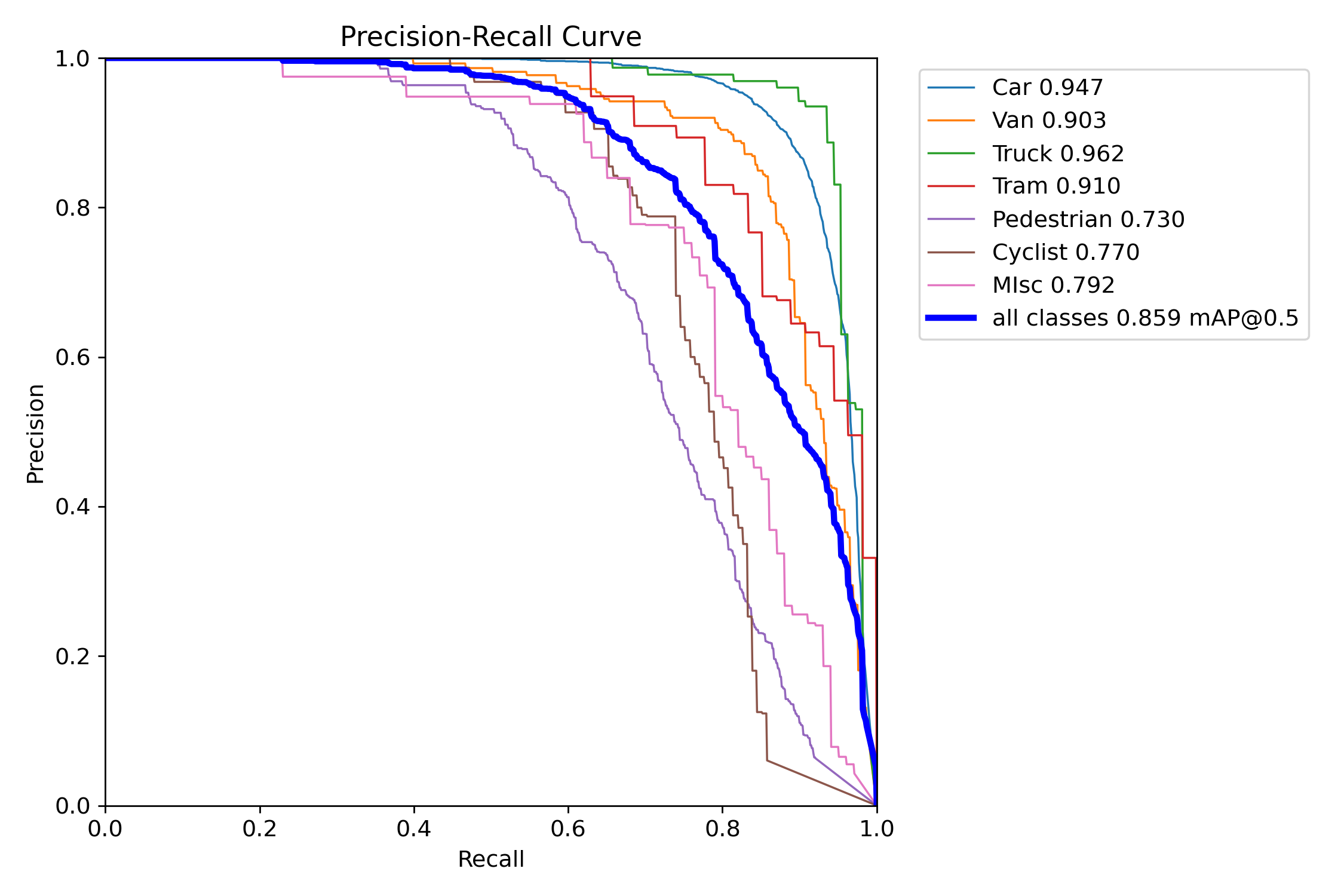}
    \caption{}
    \label{fig8-2}
  \end{subfigure}
  \hfill
  \begin{subfigure}{0.48\textwidth}
    \centering
    \includegraphics[width=\textwidth]{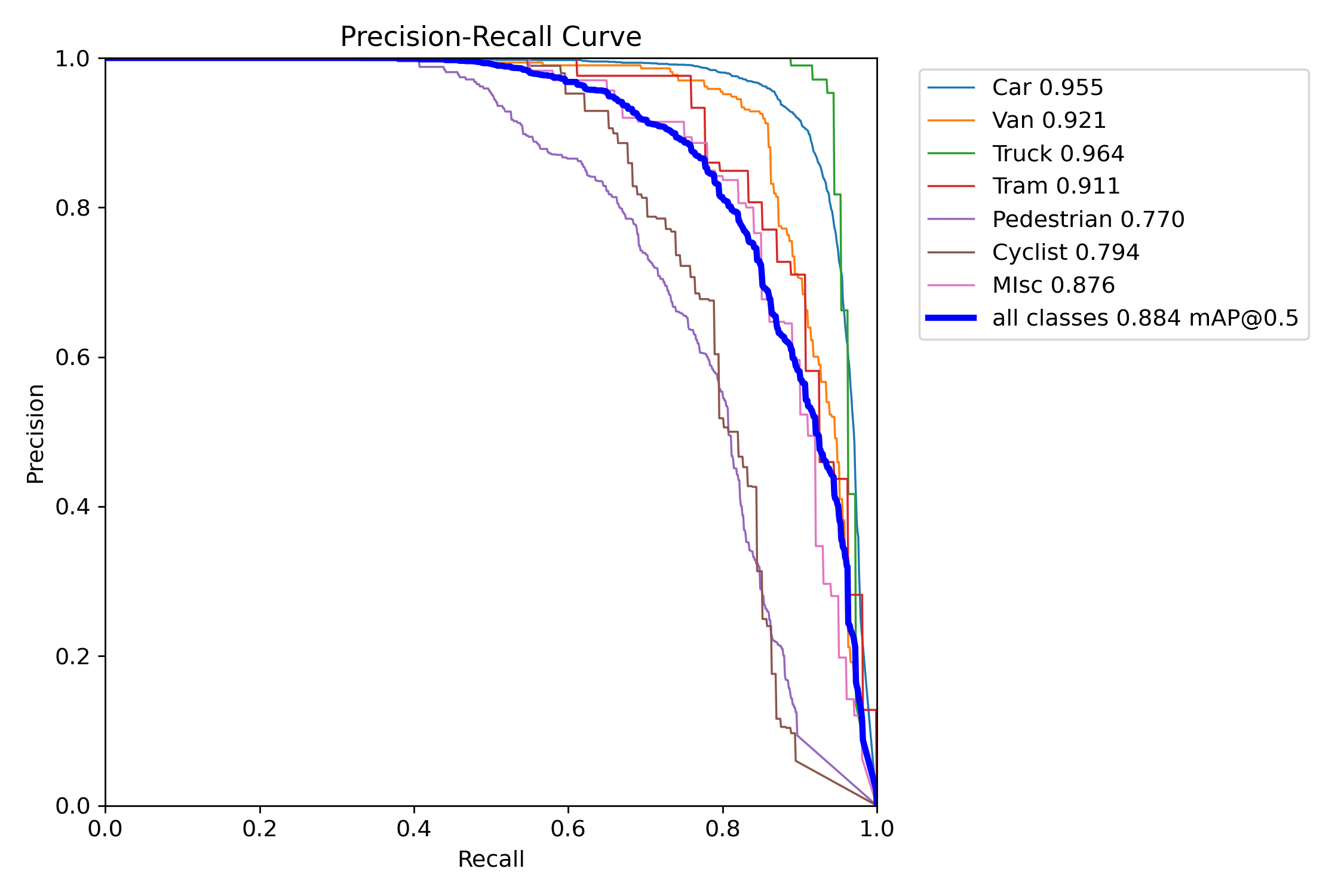}
    \caption{}
    \label{fig8-3}
  \end{subfigure}
  \hfill
  \begin{subfigure}{0.48\textwidth}
    \centering
    \includegraphics[width=\textwidth]{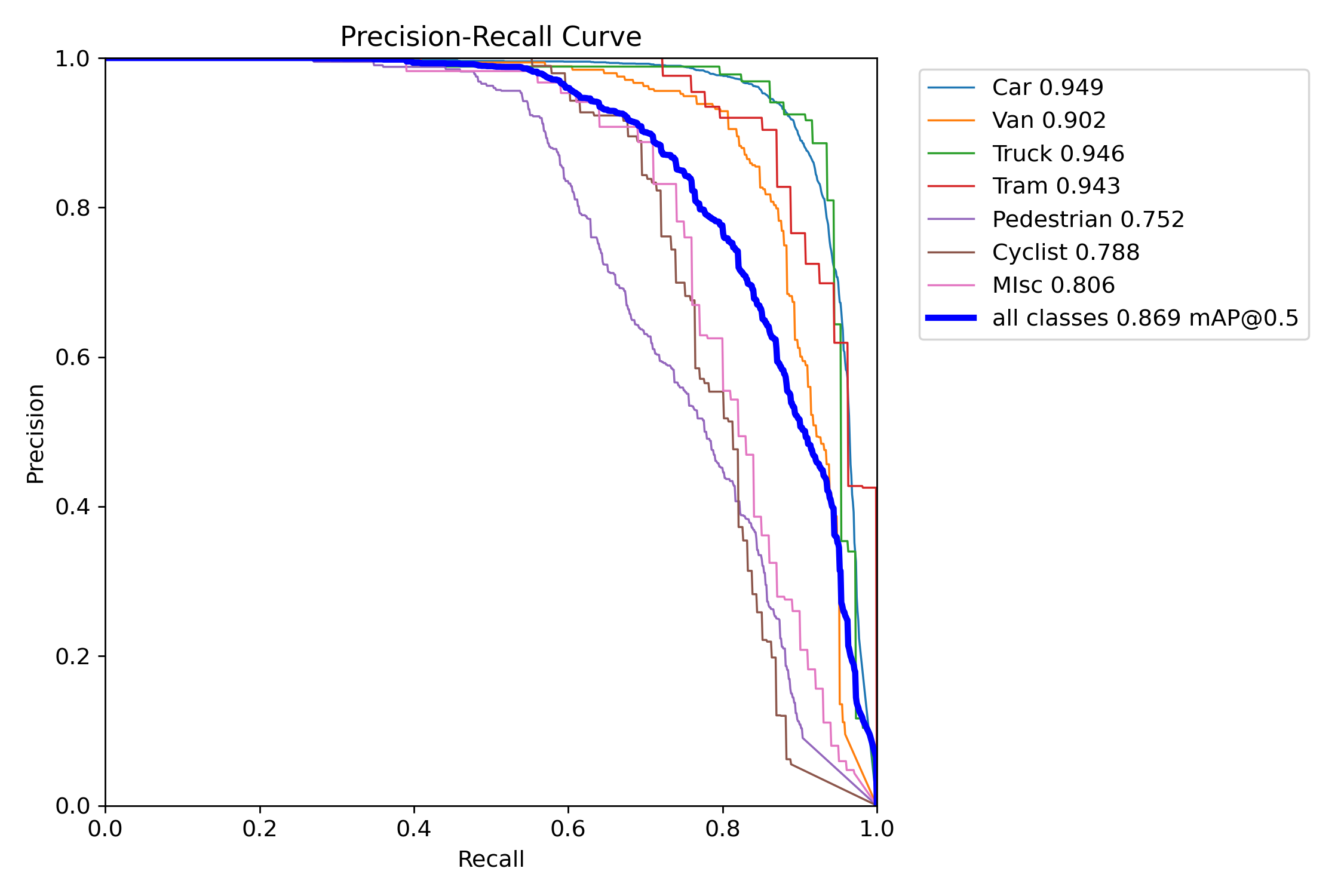}
    \caption{}
    \label{fig8-4}
  \end{subfigure}
  \hfill
  \begin{subfigure}{0.48\textwidth}
    \centering
    \includegraphics[width=\textwidth]{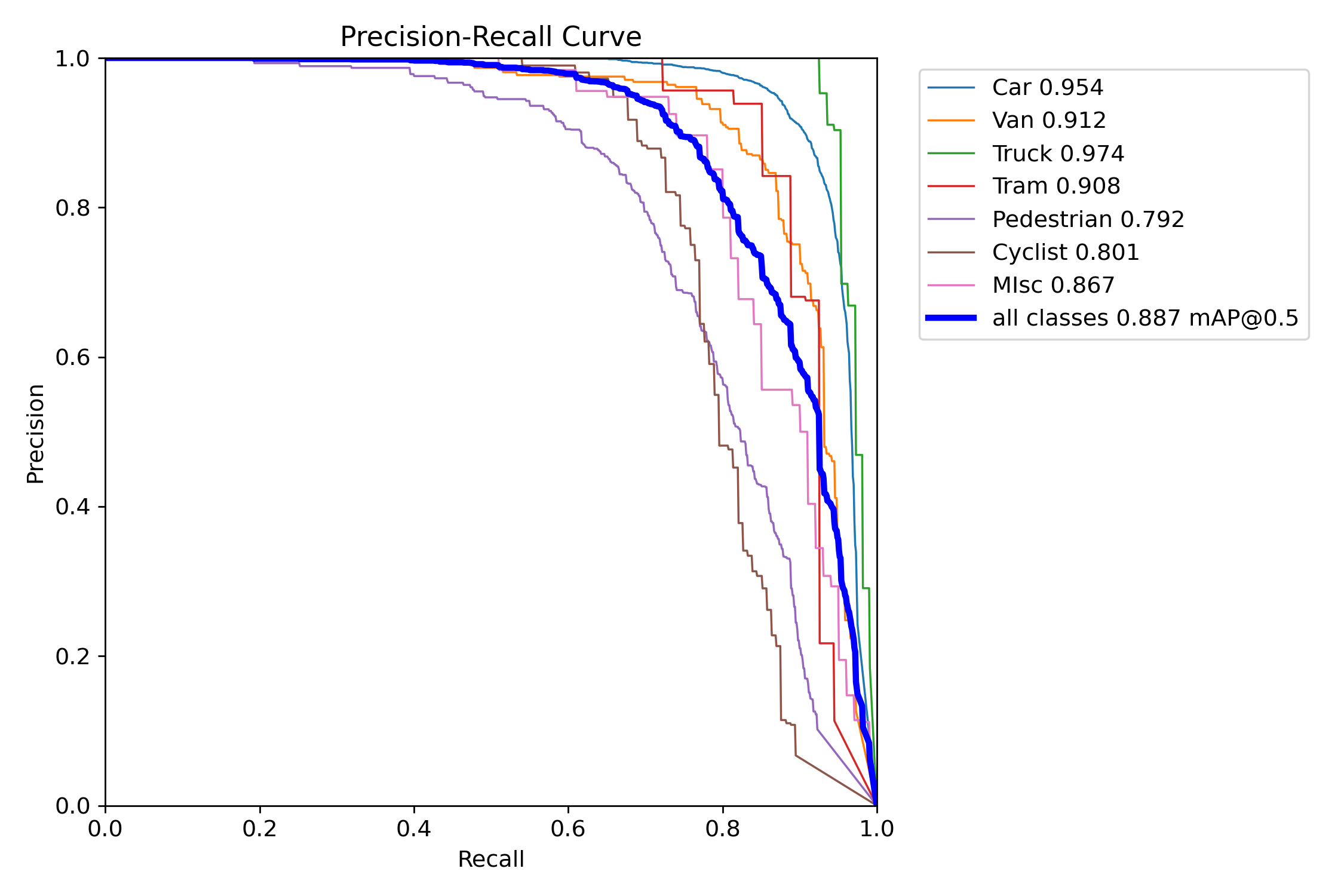}
    \caption{}
    \label{fig8-5}
  \end{subfigure}
  \hfill
  \begin{subfigure}{0.48\textwidth}
    \centering
    \includegraphics[width=\textwidth]{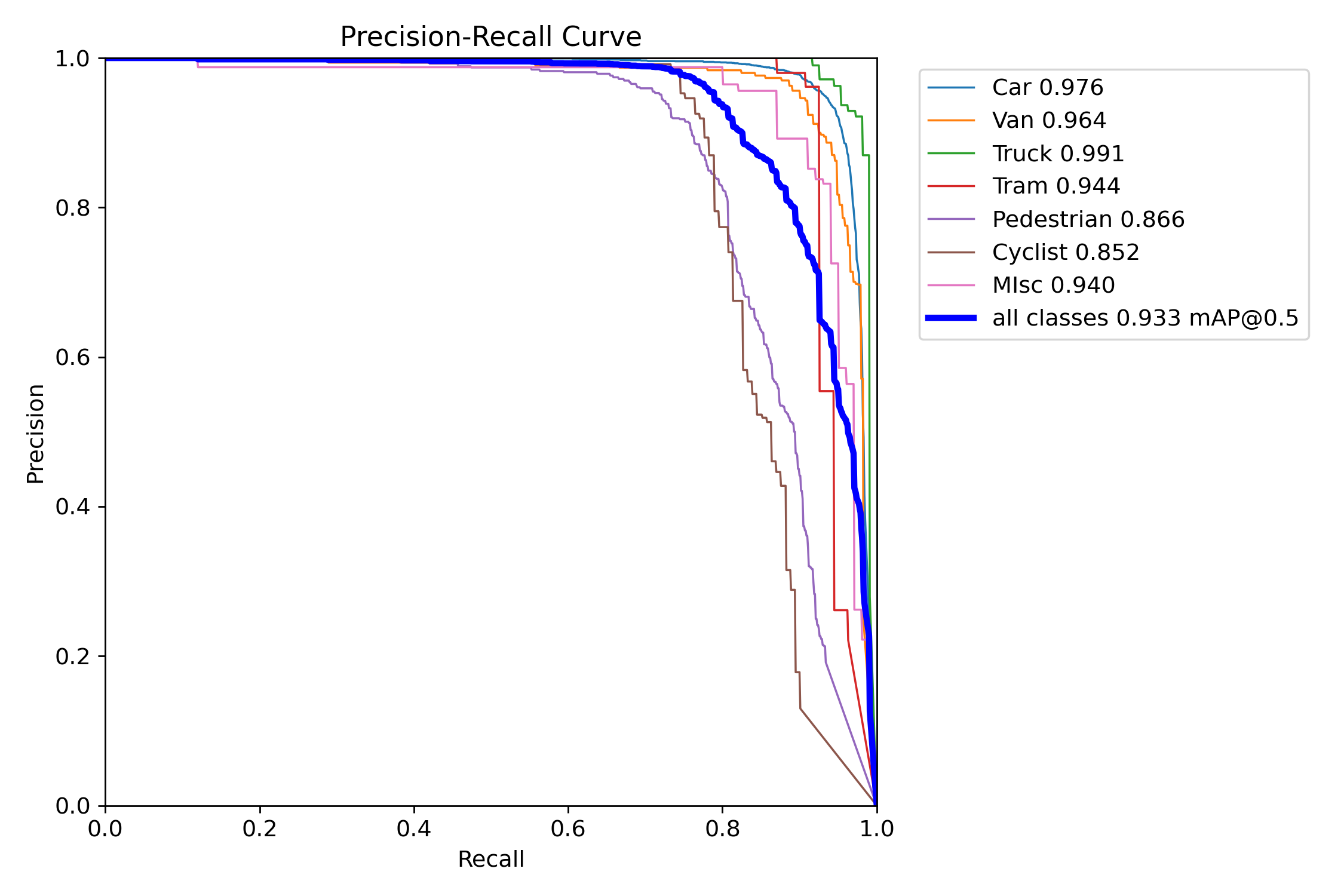}
    \caption{}
    \label{fig8-6}
  \end{subfigure}
  \caption{PR curves of YOLOv8n (a), YOLOv10n (b), YOLOv11n (c), YOLOv12n (d), YOLOv13n (e) and our MDDCNet-T (f) on the KITTI dataset.}
  \label{fig:comp_PR_KITTI}
\end{figure*}

Fig.~\ref{fig:comp_PR_KITTI} presents Precision-Recall (PR) curves of YOLOv8n, YOLOv10n, YOLOv11n, YOLOv12n, YOLOv13n, and our MDDCNet-T on the KITTI dataset for comparison. It can be seen that the MDDCNet achieves the highest AUC (Area Under the Curve), while maintaining more stable detection performance across different confidence thresholds. In contrast, some YOLO models demonstrate a significant decrease in precision at high recall rates, which tends to introduce more false positives. This result indicates that our MDDCNet enjoys stronger robustness in discriminating object confidence and distinguishing between positive and negative samples, thus effectively improving the overall detection reliability.

In addition to the evaluations on the public KITTI dataset, additional experiments are carried out on our self-constructed RTOD dataset to evaluate the practicability of our MDDCNet. Table~\ref{Tab:comprehensive_RTOD} presents comprehensive comparison of our method and different lightweight object detection models on the RTOD dataset. It can be observed that our efficient MDDCNet-T achieves superior performance on the RTOD dataset, reporting the highest 85.3\% mAP@50 and 61.6\% mAP@50-95 with 2$\times$$\sim$3$\times$ inference speed of mainstream YOLO detectors. Furthermore, our MDDCNet-T not only beats the other competing detectors with superior comprehensive performance but excels in class-specific detection accuracy as shown in Table~\ref{Tab:class_RTOD}. These results sufficiently demonstrate the promising potential of our MDDCNet framework in achieving preferable tradeoff between accuracy and efficiency for practical applications.

\begin{table*}
  \centering
  \caption{Comprehensive comparison of different models on the RTOD Dataset.}
  \label{Tab:comprehensive_RTOD}
  \resizebox{\linewidth}{!}{
  \begin{tabular}{l|c|c|c|c|c}
    \hline
    \textbf{Detectors} & \textbf{P(\%)} & \textbf{R(\%)} & \textbf{mAP@50(\%)} & \textbf{mAP@50-95(\%)} & \textbf{FPS} \\
    \hline
    YOLOv5n & 87.5 & 76.1 & 82.1 & 57.6 & 46.18 \\
    \hline
    YOLOv6s & 84.3 & 73.3 & 79.2 & 55.2 & 48.82 \\
    \hline
    YOLOv8n & 89.9 & 79.1 & 84.3 & 60.6 & 55.56 \\
    \hline
    YOLOv9t & 83.9 & 77.3 & 81.2 & 57.9 & 51.28 \\
    \hline
    YOLOv11n & 87.1 & 78.9 & 84.3 & 60.1 & 55.25 \\
    \hline
    YOLOv12n & 89.2 & 77.4 & 81.7 & 58.0 & 52.36 \\
    \hline
    YOLOv13n & 85.8 & 71.8 & 80.0 & 55.9 & 54.35 \\
    \hline
    MambaYOLO-T & 88.3 & 77.2 & 82.5 & 60.4 & 42.73 \\
    \hline
    \hline
    \textbf{MDDCNet-N} & \textbf{88.5} & \textbf{78.2} & \textbf{82.4} & \textbf{61.6} & \textbf{52.08} \\
    \textbf{MDDCNet-T} & \textbf{90.2} & \textbf{77.5} & \textbf{85.3} & \textbf{61.6} & \textbf{45.06} \\
    \hline
  \end{tabular}
  }
\end{table*}

\begin{figure}
  \centering
  \begin{subfigure}{0.48\textwidth}
    \centering
    \includegraphics[width=\textwidth]{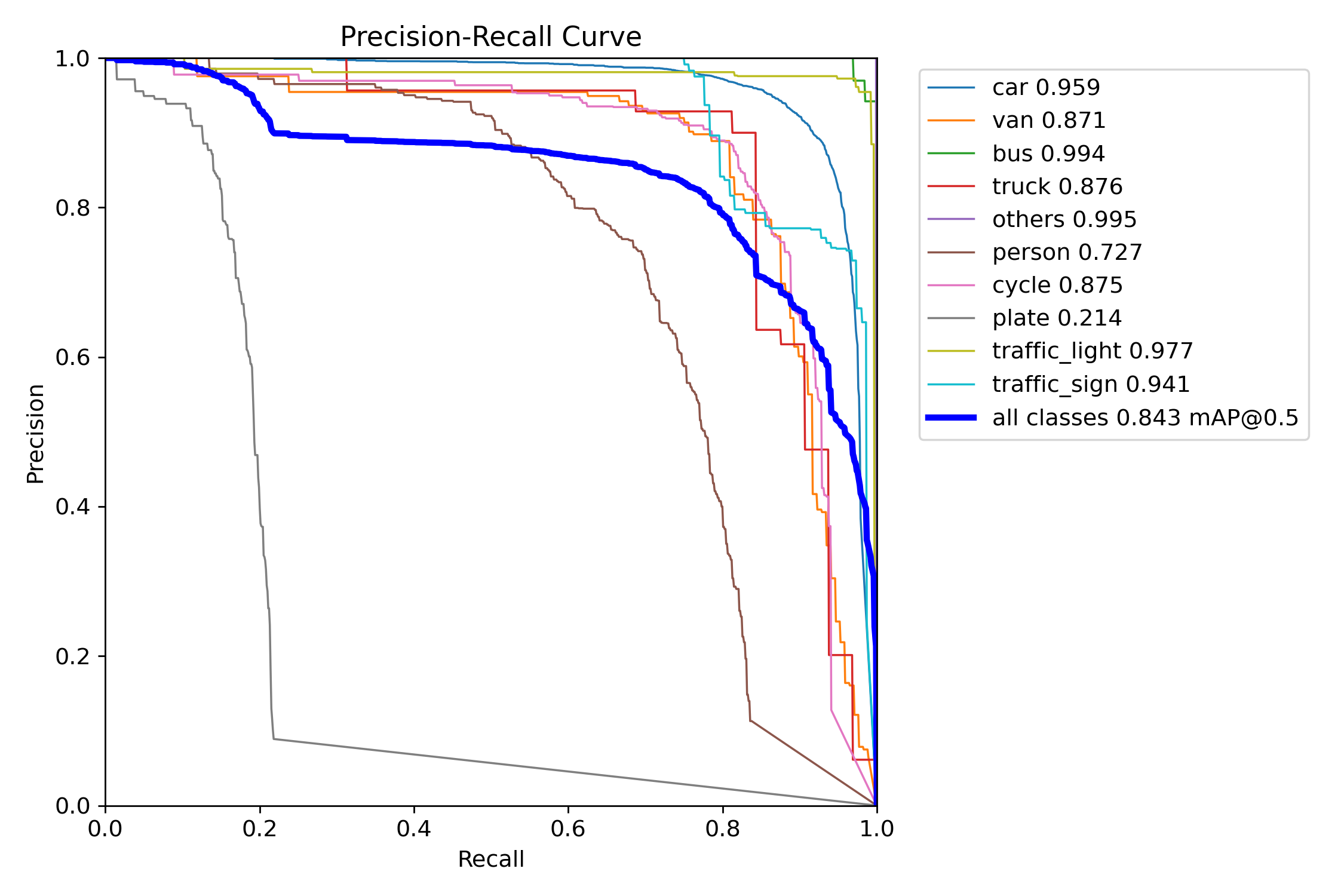}
    \caption{}  
    \label{fig9-1}
  \end{subfigure}
  \hfill
  \begin{subfigure}{0.48\textwidth}
    \centering
    \includegraphics[width=\textwidth]{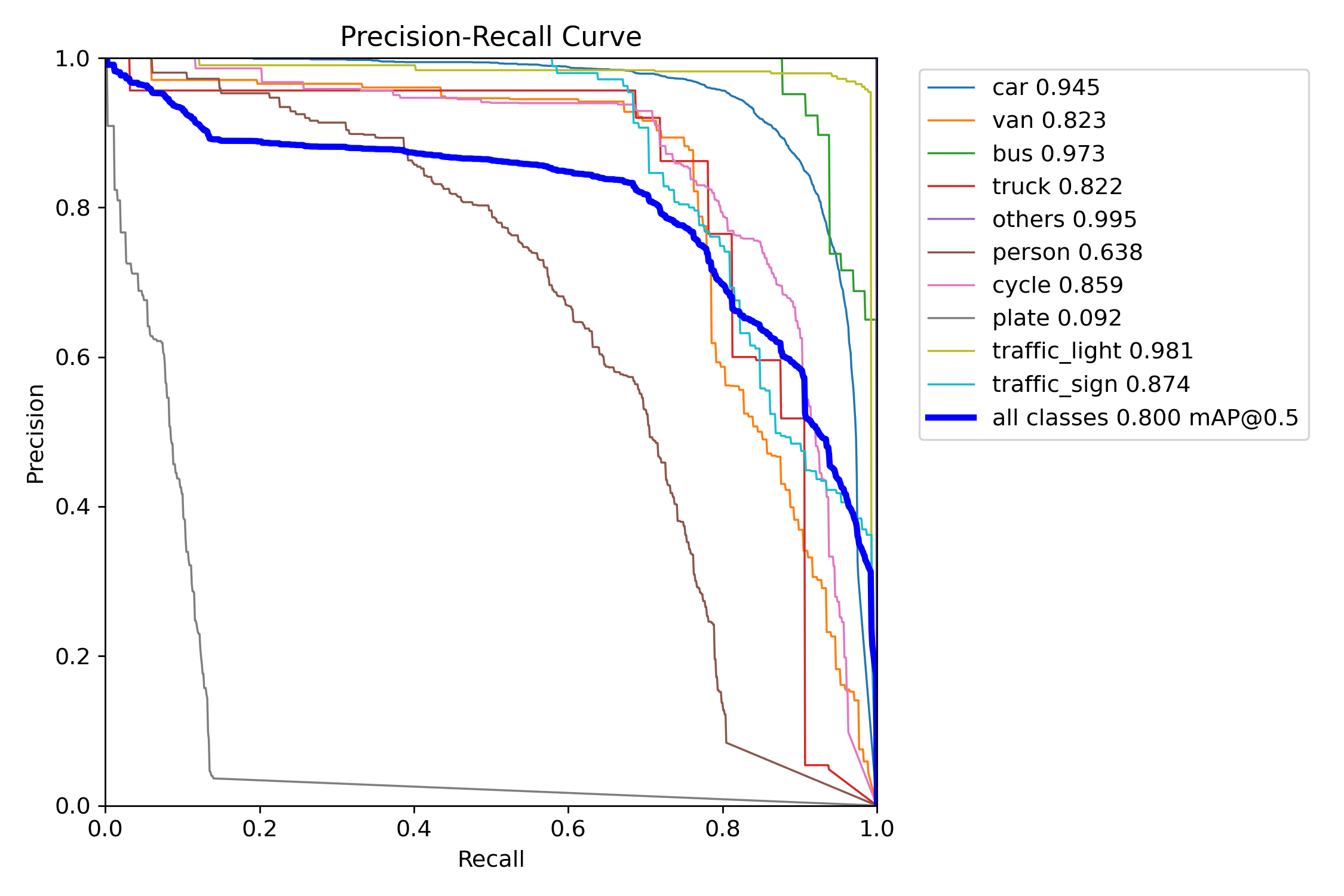}
    \caption{}
    \label{fig9-2}
  \end{subfigure}
  \hfill
  \begin{subfigure}{0.48\textwidth}
    \centering
    \includegraphics[width=\textwidth]{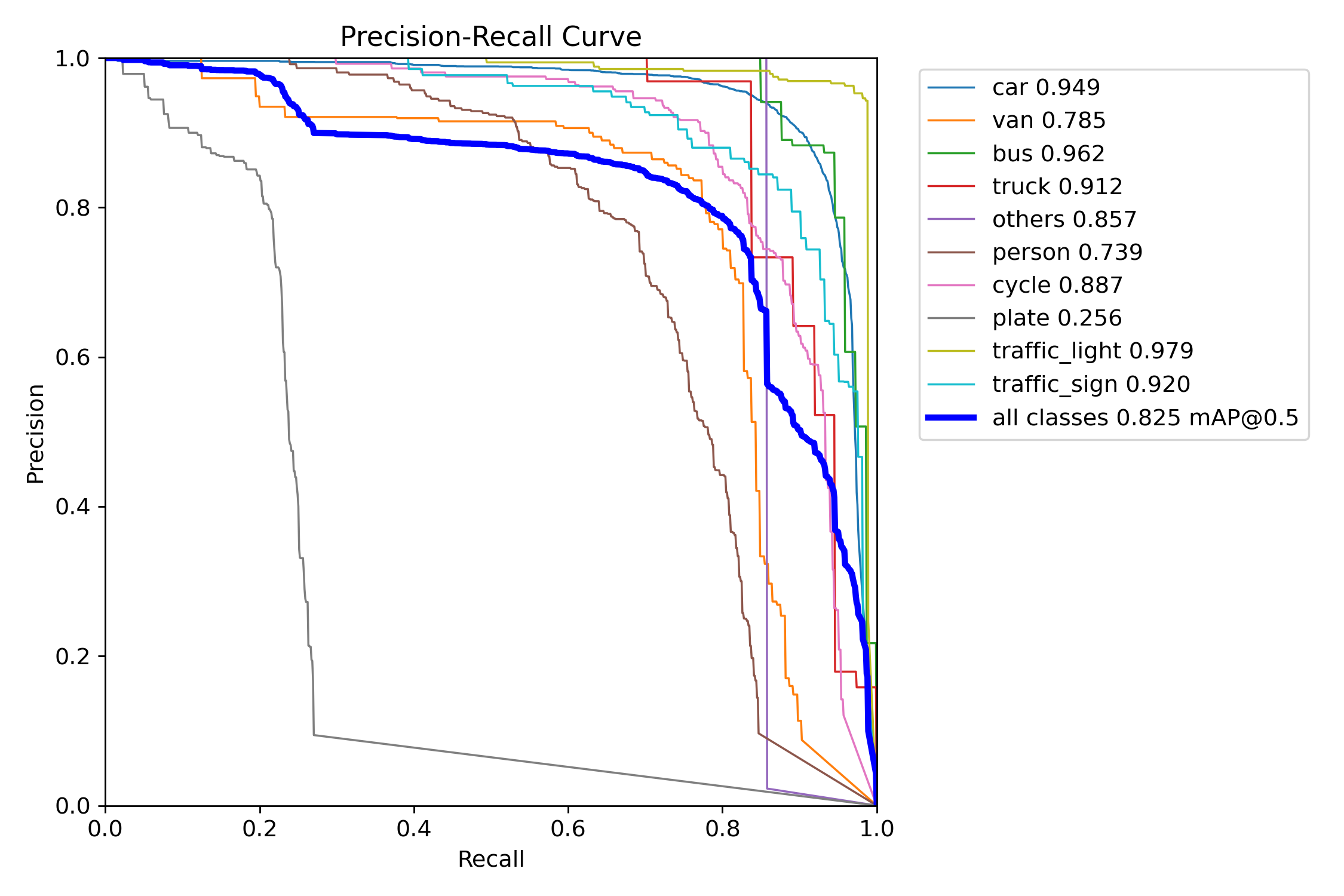}
    \caption{}
    \label{fig9-3}
  \end{subfigure}
  \hfill
  \begin{subfigure}{0.48\textwidth}
    \centering
    \includegraphics[width=\textwidth]{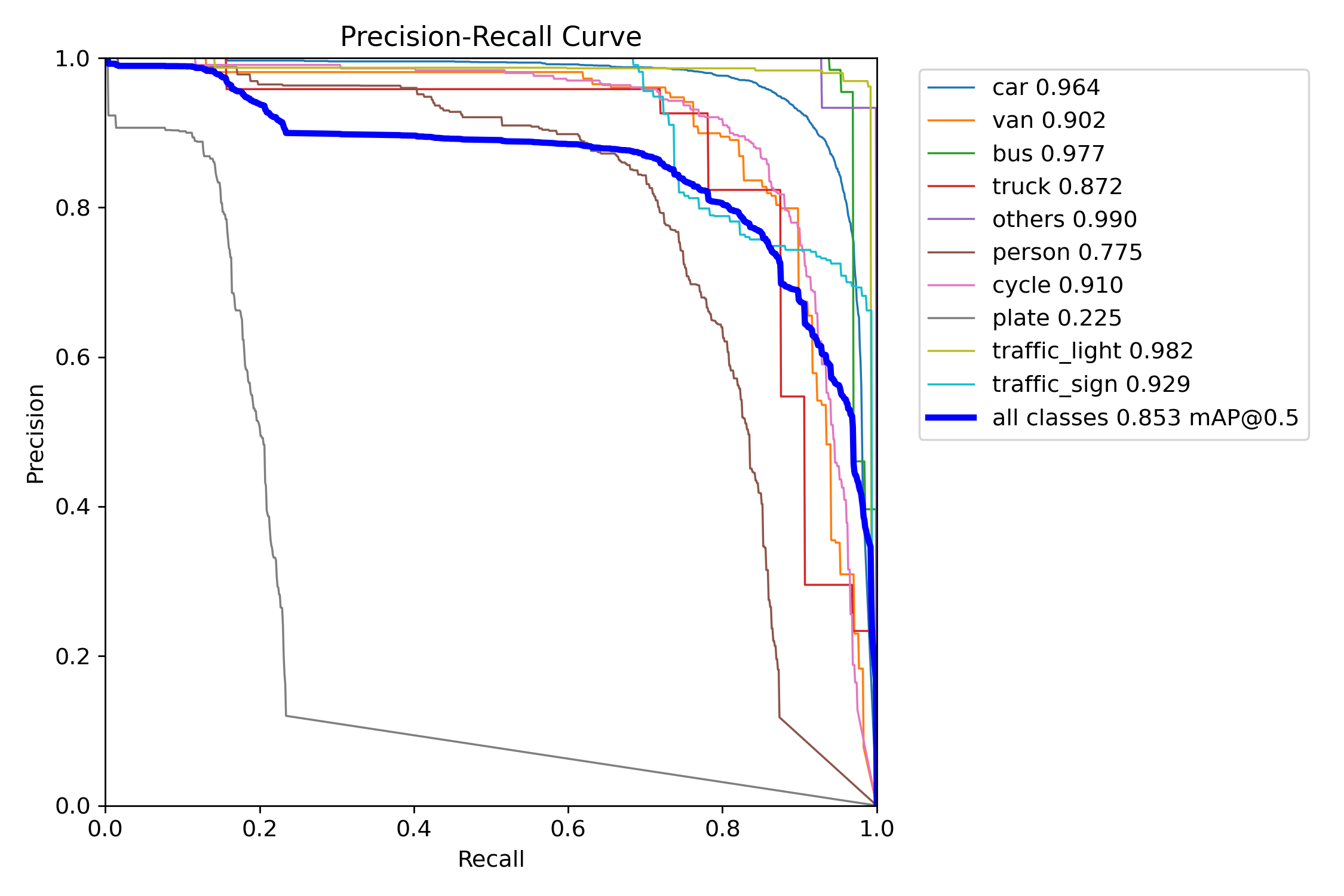}
    \caption{}
    \label{fig9-4}
  \end{subfigure}
  \caption{PR curves of YOLOv8n (a), YOLOv13n (b), MambaYOLO-T (c), and our MDDCNet-T (d) on the RTOD dataset.}
  \label{fig:comp_PR_RTOD}
\end{figure}

Fig.~\ref{fig:comp_PR_RTOD} presents a comparison of PR curves of four different models on the RTOD dataset. Consistent with the results on KITTI dataset, the PR curve of our MDDCNet is generally smoother and maintains relatively high Precision even at high recall rates. This verifies the advantages of our method in effectively reducing both false positives and missed detections. In contrast, the precision of the other lightweight models declines more rapidly as recall increases, indicating that they fail to distinguish between positive and negative samples in complex traffic scenarios.

\begin{table*}
  \centering
  \caption{Comparison of different detectors across various categories on the RTOD dataset (mAP@50\%).}
  \label{Tab:class_RTOD}
  \resizebox{\linewidth}{!}{
  \begin{tabular}{l|c|c|c|c|c|c|c|c|c|c}
    \hline
    \textbf{Detectors} & \textbf{Car} & \textbf{Van} & \textbf{Bus} & \textbf{Truck} & \textbf{Person} & \textbf{Cycle} & \textbf{Plate} & \textbf{Traffic light} & \textbf{Traffic sign} & \textbf{Others} \\
    \hline
    YOLOv5n & 95.7 & 87.2 & 96.9 & 84.2 & 71.0 & 89.1 & 10.0 & 98.3 & 94.5 & 94.1 \\
    \hline
    YOLOv6s & 94.9 & 79.9 & 97.8 & 78.5 & 67.1 & 89.8 & 2.1 & 97.7 & 94.1 & 89.8 \\
    \hline
    YOLOv8n & 95.9 & 87.1 & 99.4 & 87.6 & 72.7 & 87.5 & 21.4 & 97.7 & 94.1 & 99.5\\
    \hline
    YOLOv9t & 95.6 & 87.1 & 98.6 & 82.6 & 67.4 & 89.9 & 8.6 & 97.9 & 88.3 & 95.9 \\
    \hline
    YOLOv11n & 95.6 & 86.2 & 96.8 & 88.7 & 71.9 & 91.1 & 23.3 & 97.9 & 93.5 & 97.4 \\
    \hline
    YOLOv12n & 95.4 & 86.9 & 97.7 & 86.0 & 70.8 & 90.0 & 0.3 & 98.0 & 94.9 & 96.7 \\
    \hline
    YOLOv13n & 94.5 & 82.3 & 97.3 & 82.2 & 63.8 & 85.9 & 9.2 & 98.1 & 87.4 & 99.5 \\
    \hline
    MambaYOLO-T & 94.9 & 78.5 & 96.2 & 91.2 & 73.9 & 88.7 & 25.6 & 97.9 & 92.0 & 85.7 \\
    \hline
    \hline
    \textbf{MDDCNet-N} & \textbf{94.8} & \textbf{79.4} & \textbf{96.4} & \textbf{96.5} & \textbf{73.0} & \textbf{88.0} & \textbf{18.5} & \textbf{97.4} & \textbf{94.6} & \textbf{85.7} \\
    \textbf{MDDCNet-T} & \textbf{96.4} & \textbf{90.2} & \textbf{97.7} & \textbf{87.2} & \textbf{77.5} & \textbf{91.0} & \textbf{22.5} & \textbf{98.2} & \textbf{92.9} & \textbf{99.0} \\
    \hline
  \end{tabular}
  }
\end{table*}

\begin{table*}
  \centering
  \caption{Ablation experiments on backbone structures.}
  \label{Tab:abl_backbone}
  \resizebox{\linewidth}{!}{
  \begin{tabular}{c|c|c|c|c}
    \hline
    \textbf{Backbone architectures} & \textbf{mAP@50(\%)} & \textbf{mAP@50-95(\%)} & \textbf{Params(M)} & \textbf{FLOPs(G)} \\
    \hline
    [MSDDC,MSDDC,MSDDC,MSDDC] & 91.6 & 68.4 & 7.4 & 13.4 \\
    \hline
    [Mamba,Mamba,Mamba,Mamba] & 91.5 & 69.7 & 4.9 & 11.1 \\
    \hline 
    [Mamba,Mamba,MSDDC,MSDDC] & 91.7 & 69.1 & 7.3 & 12.9 \\
    \hline
    \textbf{[MSDDC,MSDDC,Mamba,Mamba]} & \textbf{92.1} & \textbf{69.7} & \textbf{5.0} & \textbf{11.6} \\
    \hline
  \end{tabular}
  }
\end{table*}

\subsection{Ablation experiments}
In this section, we have conducted extensive ablation studies to explore the effect of key modules on our model performance. Notably, \textbf{all the ablation experiments are carried out on the KITTI dataset.}

\paragraph{Backbone network}
To explore the efficacy of different building blocks within our hybrid Backbone, we design and validate four different hybrid architectures, i.e., a backbone with only MSDDC blocks, a backbone with only Mamba blocks, and two hybrid variants with different module sequences. As shown in Table~\ref{Tab:abl_backbone}, our designed [MSDDC, MSDDC, Mamba, Mamba] backbone brings the highest 92.1\% mAP@50 and 69.7\% mAP@50-95 accuracies with only 5M parameters. These results verify that the complementary advantages between local cues and global semantics can be achieved by capturing local details features (e.g., vehicle edges and textures) via MSDDC Blocks in the shallow stages, followed by employing Mamba Block in the deeper stages to model long-range spatial dependencies (e.g., distant small targets and road structural context). This highly resembles the inherent human visual perception process in which the capture of low-level features typically precedes the generation of high-level semantics. In contrast, swapping the order of MSDDC and Mamba blocks yields inferior performance, which fully suggests that the perceptual sequence plays a crucial role in accurate object recognition and detection. 

\begin{table*}
  \centering
  \caption{Ablation experiments on the Dilation Rates (DR) in our MSDDC module.}
  \label{Tab:abl_DR}
  \resizebox{\linewidth}{!}{
  \begin{tabular}{c|c|c|c|c|c|c}
    \hline
    \textbf{Setting} & \textbf{P(\%)} & \textbf{R(\%)} & \textbf{mAP@50(\%)} & \textbf{mAP@50-95(\%)} & \textbf{Params(M)} & \textbf{FLOPs(G)} \\
    \hline
    DR=1,2,3 & 92.2 & 84.7 & 91.6 & 69.9 & 5.0 & 11.6 \\
    \hline
    DR=1,3,5 & 91.5 & 84.8 & 91.9 & 68.6 & 5.0 & 11.6 \\
    \hline
    \textbf{DR=1,2,4} & \textbf{93.4} & 83.9 & \textbf{92.1} & \textbf{69.7} & \textbf{5.0} & \textbf{11.6} \\
    \hline
  \end{tabular}
  }  
\end{table*}

\paragraph{MSDDC module}
As shown in Table~\ref{Tab:abl_DR}, we explore the impact of different Dilation Rate (DR) settings within our MSDDC module on our model performance. It can be clearly observed that the setting of {1, 2, 4} achieves the highest 92.1\% mAP@50, while maintaining a compact network and unchanged computational complexity. The results reveal that a more balanced receptive field expansion across different scales and enhanced multi-scale feature representation capability can be obtained with a reasonable setting of DR combination. Therefore, the proposed structure can simultaneously improve detection accuracy and inference efficiency without increasing computational overhead, providing an effective solution for real-time traffic object detection.

\begin{table}
  \centering
  \caption{Ablation experiments on different attention mechanisms.}
  \label{Tab:abl_attention}
  \resizebox{\textwidth}{!}{
  \begin{tabular}{c|c|c|c|c|c|c}
    \hline
    \textbf{Attention} & \textbf{P(\%)} & \textbf{R(\%)} & \textbf{mAP@50(\%)} & \textbf{mAP@50-95(\%)} & \textbf{Params(M)} & \textbf{FLOPs(G)} \\
    \hline
    Concat & 92.2 & 84.7 & 91.5 & 69.7 & 4.9 & 11.1 \\
    \hline
    MLCA-only & 91.6 & 83.4 & 91.2 & 69.4 & 5.0 & 11.3 \\
    \hline 
    RCSSC & 92.8 & 86.4 & 91.8 & 71.3 & 5.3 & 12.0 \\
    \hline
    \textbf{CSCA} & \textbf{93.9} & 85.4 & \textbf{92.5} & \textbf{71.1} & 5.5 & 11.6 \\
    \hline
  \end{tabular}
  }
\end{table}

\begin{table}
    \centering
    \caption{Ablation experiments on different FFN structures.} 
    \begin{tabular}{c|c|c|c|c|c|c}
    \hline
       \textbf{Methods} & \textbf{P(\%)} & \textbf{R(\%)} & \textbf{mAP@50(\%)} & \textbf{mAP@50-95(\%)} & \textbf{Params(M)} & \textbf{FLOPs(G)} \\
       \hline
       Vanilla FFN & 92.2 & 84.7 & 91.5 & 69.7 & 4.9 & 11.1 \\
       \hline
       CA Block & 92.0 & 84.2 & 91.7 & 69.9 & 5.3 & 11.5 \\
       \hline
       Residual CA Block & 91.0 & 85.7 & 91.8 & 69.4 & 5.3 & 11.6 \\
       \hline
       Gated CA Block & 91.2 & 85.3 & 92.0 & 69.6 & 5.5 & 11.7 \\
       \hline
       CE-FFN & 94.2 & 84.0 & 92.3 & 70.0 & 5.7 & 11.8 \\
       \hline
    \end{tabular}
    \label{tab:abl_FFN}
\end{table}

\paragraph{CE-FFN module}
Additionally, we compare our designed CE-FFN with other FFN structures as presented in Table~\ref{tab:abl_FFN}. It is clearly shown that our CE-FFN exceeds the others by reporting the highest 92.3\% mAP@50 and 70.0\% mAP@50-95, which suggests that CE-FFN not only inherits the strong representation capability from CA block via channel attention but also employs a complementary global branch to further strengthen feature interaction, thereby enabling local-global collaborative modeling is achieved to boost detection accuracy. Although residual CA block and gated CA block are capable of capturing local details with enhanced representation, they only extend the vanilla CA block with residual connections or gating mechanisms without introducing explicit global context modeling, thus exhibiting inferior performance compared to our CE-FFN.

\paragraph{CSCA module}
To delve into the impact of attention mechanism in FPN, we have systematically compared four attention strategies, namely direct attention concatenation, MLCA-only, RCSSC\cite{RCSSC}, and our CSCA module. As shown in Table~\ref{Tab:abl_attention}, our proposed CSCA achieves the best performance of 92.5\% mAP@50, which consistently the other attention strategies. This result demonstrates that CSCA can comprehensively capture discriminative spatial-wise and channel-wise details as well as contextual information, enabling adaptive synergy of multi-scale features and substantially benefits cross-scale feature interaction. In contrast, the other attention strategies fail to sufficiently model comprehensive feature interactions, thereby exhibiting degraded performance. For example, MLCA-only attention overlooks discriminative spatial information and cannot capture multi-scale contextual dependencies, leading to insufficient feature fusion and limited detection performance.

\begin{table*}[tbp]
  \centering
  \caption{Comprehensive ablation experiments of our MDDCNet model.}
  \label{Tab:abl_comprehensive}
  \resizebox{\linewidth}{!}{
  \begin{tabular}{c|c|c|c|c|c|c|c|c|c}
    \hline
    \multicolumn{3}{c|}{\textbf{Modules}} & \multirow{2}{*}{\textbf{P(\%)}} & \multirow{2}{*}{\textbf{R(\%)}} & \multirow{2}{*}{\textbf{mAP@50(\%)}} & \multirow{2}{*}{\textbf{mAP@50-95(\%)}} & \multirow{2}{*}{\textbf{Params(M)}} & \multirow{2}{*}{\textbf{FLOPs(G)}} \\
    \cline{1-3}
    \textbf{CE-FFN} & \textbf{MSDDC} & \textbf{CSCA} & & & & & & \\
    \hline
    & & & 92.5 & 84.3 & 91.5 & 69.7 & 4.9 & 11.1  \\
    \hline
    \checkmark & & & 94.2 & 84.0 & 92.3 & 70.0 & 5.7 & 11.8  \\
    \hline
     & \checkmark & & 93.4 & 83.9 & 92.1 & 69.7 & 5.0 & 11.6 \\
    \hline
     & & \checkmark & 93.9 & 85.4 & 92.5 & 71.1& 5.5 & 11.6 \\
    \hline
    \checkmark & \checkmark &  & 92.2 & 85.8 & 92.8 & 69.3 & 6.0 & 12.2 \\
    \hline
    \checkmark & \checkmark & \checkmark & \textbf{95.3} & \textbf{86.1} & \textbf{93.3} & \textbf{72.3} & 6.6 & 12.9 \\
    \hline
  \end{tabular}
  }
\end{table*}

\paragraph{Comprehensive ablation studies}
In addition to the above-mentioned experiments, we further explore the individual influence of the three modules (MSDDC, CE-FFN, CSCA) on the model performance to provide a comprehensive insight. Utilizing our MDDCNet architecture excluding any aforementioned module as the baseline, each module is progressively incorporated to construct different models. As demonstrated in Table~\ref{Tab:abl_comprehensive}, introducing MSDDC module into the baseline increases mAP@50 from 91.5\% to 92.1\%, showcasing the benefit of MSDDC in handling scale variances and appearance deformation in complex traffic scenes. When the CE-FFN module is embedded into the backbone, a 0.8\% mAP@50 improvement can be observed, which implies that CE-FFN is capable of strengthening cross-channel interaction and thus significantly contributes to object detection in complex scenarios. Moreover, integrating CSCA module is also conducive to accurate detection, suggesting that 1\% performance boost can be obtained by adaptively fusing contextual-spatial-channel attention enhanced features. In particular, the complete model achieves the highest detection accuracy with only slight increase in computational overhead. This verifies the effectiveness of each module's design and demonstrates the significant advantage of synergistic optimization for object detection in traffic scenes.

\begin{figure*}
    \centering
    \includegraphics[width=1.0\textwidth]{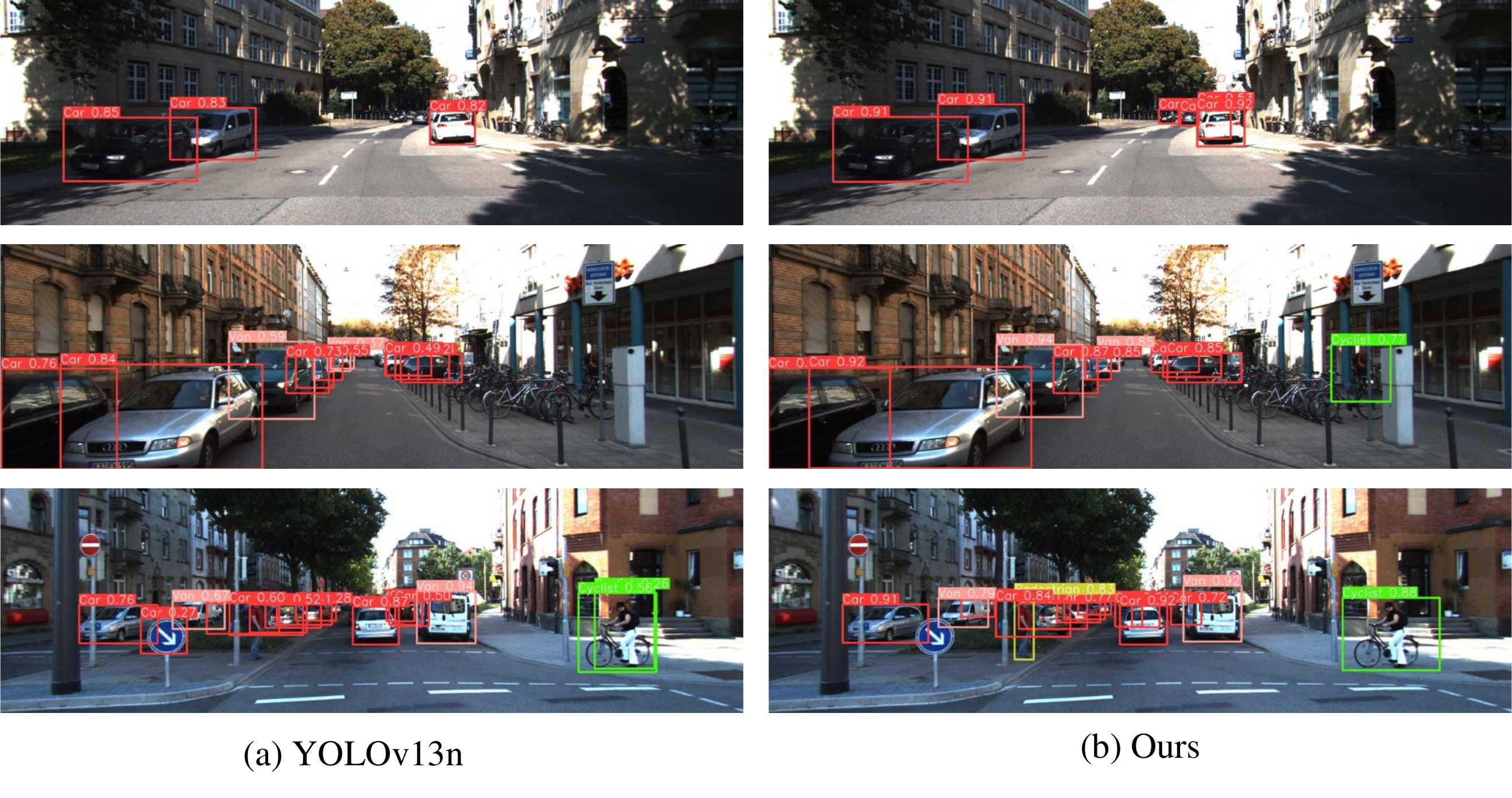}
    \caption{Qualitative comparison of YOLOv13n and our MDDCNet on the KITTI dataset. Our MDDCNet demonstrates superior performance with fewer missed detections.}
    \label{fig:qualComp_KITTI}
\end{figure*}

\begin{figure*}
    \centering
    \includegraphics[width=1.0\textwidth]{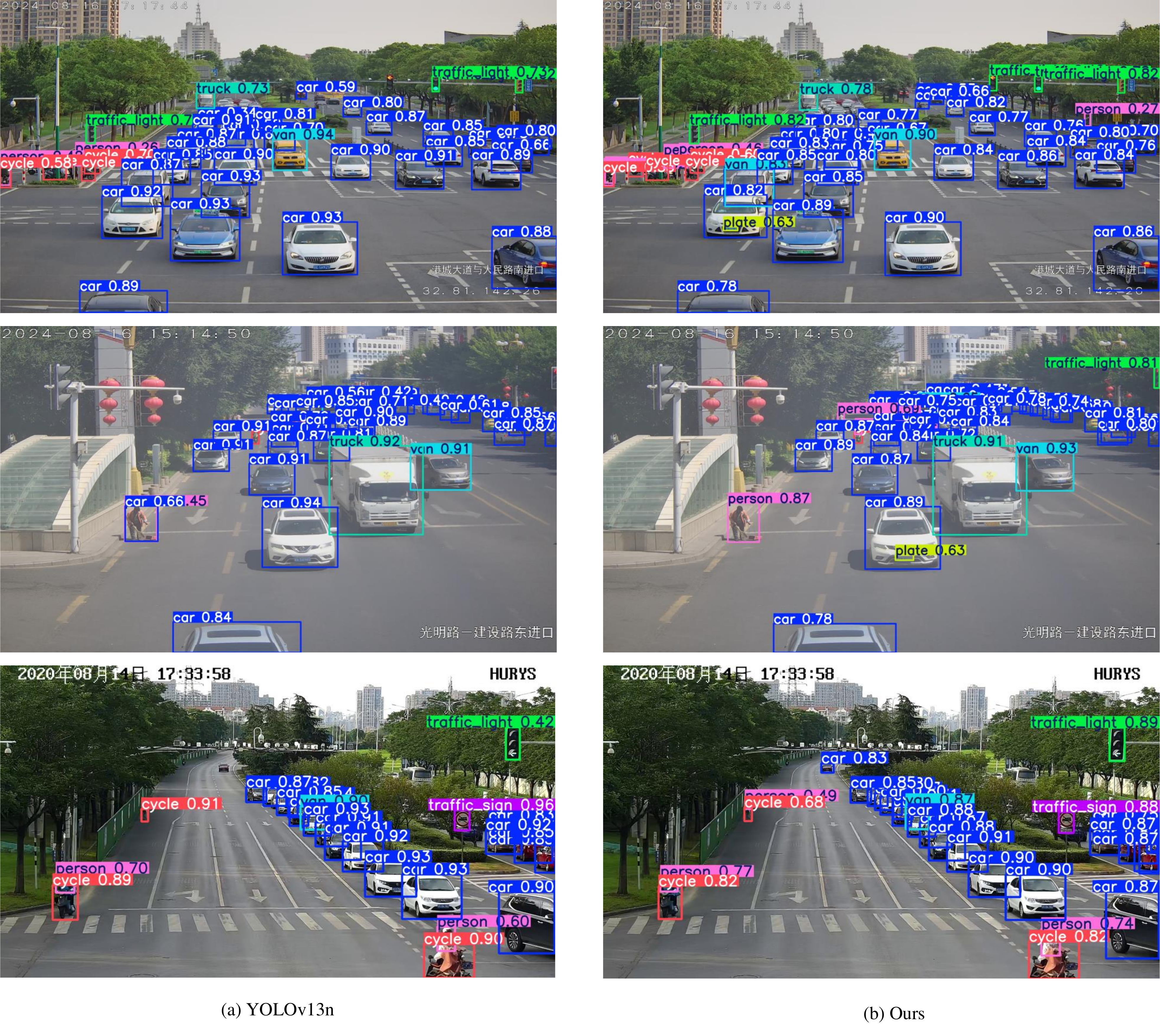}
    \caption{Qualitative comparison of YOLOv13n and our MDDCNet on the RTOD dataset. It can be observed that our method outperforms YOLOv13n by achieving more accurate results with fewer missed targets and lower false positive rates.}
    \label{fig:qualComp_RTOD}
\end{figure*}

\subsection{Qualitative evaluations}
To intuitively demonstrate the superiority of the proposed method, we have qualitatively compared our MDDCNet with the latest YOLOv13n detector on both datasets. As illustrated in Fig.~\ref{fig:qualComp_KITTI}, our MDDCNet outperforms YOLOv13n by achieving more accurate results with fewer missed detections. Specifically, YOLOv13n fails to recognize the distant cars, the partially occluded cyclist and the standing pedestrian, whereas our MDDCNet accurately detects these targets with high confidence. On the real-world RTOD dataset, the advantage of our method is sufficiently manifested in accurate cross-scale object detection with better robustness as shown in Fig.~\ref{fig:qualComp_RTOD}. For example, in a lightly foggy scenario (the second row), MDDCNet is capable of accurately identifying multi-scale objects ranging from close-range person and car plate to distant walking pedestrian. In particular, even the partial occluded traffic light can still be detected by our model with a high confidence score. In contrast, YOLOv13n misses the traffic light and car plate, whilst misclassified the person as the car. In other cases, our MDDCNet also excels at accurately detecting distant tiny objects, e.g., can and person, both of which are missed in the detection results obtained by YOLOv13n.

\section{Conclusion}
In this study, we propose a Mamba with Deformable Dilated Convolutions Network (MDDCNet) for multi-scale object detection in complex traffic scenarios. Our MDDCNet features a hybrid CNN-Mamba backbone to achieve the synergy of local perception and global modeling. In particular, the Multi-Scale Deformable Dilated Convolution (MSDDC) blocks in shallow stages can capture fine-grained local details while exhibiting certain robustness to scale variations and geometric deformations. Meanwhile, the Mamba blocks in deep stages model global semantics via long-range context modeling while maintaining linear computational complexity. Within both MSDDC and Mamba blocks, our carefully designed Channel-Enhanced FeedForward Network (CE-FFN) module further strengthens inter-channel interaction. Moreover, we devise an Attention-Aggregating Feature Pyramid Network ($A^2$FPN) to facilitate effective multi-scale feature fusion and interaction. Extensive comparative and ablation experiments conducted on both public benchmarks and real-world dataset validate the advantages of our model against a wide range of state-of-the-art detectors, demonstrating the promise and the potential of our method for real-world traffic detection applications.


\section*{Acknowledgments}
This work was supported by the National Natural Science Foundation of China under Grant No. 62173186 and 62076134.





\bibliographystyle{cas-model2-names} 
\bibliography{ref}





\end{document}